\documentclass{mynle}
\usepackage{amsmath,amssymb,eurosym,lineno,xcolor}
\usepackage[pdftex]{graphicx}

\title[Effectiveness of Data-Driven Induction of Semantic Spaces ..]
      {Effectiveness of Data-Driven Induction of Semantic Spaces and Traditional Classifiers for Sarcasm Detection}
\author[Di Gangi, Lo Bosco, Pilato]
       {M\ls A\ls T\ls T\ls I\ls A \ls A\ls N\ls T\ls O\ls N\ls I\ls N\ls O \ls D\ls I \ls G\ls A\ls N\ls G\ls I\ls\\
       	Fondazione Bruno Kessler, Trento, Italy\\
       	Universit\`a degli Studi di Trento, Trento, Italy\\
       	email: digangi@fbk.eu
       	\and
		G\ls I\ls O\ls  S\ls U\ls \'E \ls L\ls O \ls B\ls O\ls S\ls C\ls O\ls\\
       	Universit\`a degli Studi di Palermo\\ 
       	Dipartimento di Matematica e Informatica, Palermo, Italy\\
       	email: giosue.lobosco@unipa.it
       	\and
        G\ls I\ls O\ls V\ls A\ls N\ls N\ls I \ls P\ls I\ls L\ls A\ls T\ls O\ls\\
       	Italian National Research Council\\
       	ICAR-CNR, Istituto di Calcolo e Reti ad alte prestazioni, Palermo, Italy\\
       	e-mail: giovanni.pilato@cnr.it\\
       }
       
\pagerange{\pageref{firstpage}--\pageref{lastpage}}

\begin{document}

\label{firstpage}
\maketitle

\begin{abstract}
Irony and sarcasm are two complex linguistic phenomena that are widely used in everyday language and especially over the social media, but they represent two serious issues for automated text understanding. Many labeled corpora have been extracted from several sources to accomplish this task, and it seems that sarcasm is conveyed in different ways for different domains. Nonetheless, very little work has been done for comparing different methods among the available corpora. Furthermore, usually, each author collects and uses their own datasets to evaluate his own method. In this paper, we show that sarcasm detection can be tackled by applying classical machine learning algorithms to input texts sub-symbolically represented in a Latent Semantic space.  The main consequence is that our studies establish both reference datasets and baselines for the sarcasm detection problem that could serve the scientific community to test newly proposed methods.
\end{abstract}


\section{Introduction}
Affective computing has raised a great deal of interest in the last years. Picard \shortcite{picard1995affective} introduced it as a computing paradigm that relates to, arises from, or influences emotions, letting computers be both more effective in assisting humans and successful in making decisions. \\

Language, as a conceptual process, plays a key role in the perception of verbal irony and sarcasm, two well-known forms of figurative language (FL) \cite{weitzel} 
Traditionally, irony as a figure of speech can be intended as ``saying something while meaning something else'' \cite{attardo10}. A comprehensive overview of different theories of irony has been illustrated in Attardo \shortcite{attardo07}.
Understanding if irony and sarcasm are the same linguistic phenomenon or not is still an unresolved question in literature \cite{sulis16}. Some authors consider irony a more general form of sarcasm, while others tend to consider it a separate linguistic issue \cite{ling16,wang13}. 
According to the theory of sarcastic irony, sarcasm and irony are very similar, but sarcasm has a specific victim who is the object of the sarcastic statement, while irony does not have such a target \cite{kreuz1989sarcastic}. 
More commonly, the noun ``sarcasm'' is understood as ``saying the opposite of what one is thinking''\footnote{from the Collins Dictionary}, usually with a negative intention. Henceforth, due to the different nuances of irony and sarcasm, and the multiple interpretations of these two concepts,  we do not differentiate between them, and, like many researchers, e.g., \cite{maynard2014cares},
we will use the term ``\emph{sarcasm}'' to refer to both verbal irony and sarcasm.

A sarcastic sentence may include features that characterize a positive sentiment, but that insinuates a negative sentiment~\cite{joshi2017,joshi2018}.
It is clear that sarcastic sentences are more difficult to process by an algorithm than non-sarcastic assertions; as a matter of fact, both the situation and the mental state of the speaker are factors that can determine a sarcastic content in a sentence.\\
A system capable of detecting sarcasm correctly would greatly improve the performance of sentiment analysis systems \cite{chia16,chia17,maynard2014cares,peled17}, especially considering the big data available nowadays due to the exponential growth of social platforms. 
Unfortunately, sarcasm detection in written texts is a difficult task even for humans \cite{wallace2014humans}.\\
Moreover, some people usually do not understand sarcasm, and there are sentences meant as being sarcastic by the author that are not recognized as such by the readers. 

We focus our attention on the possibility of detecting sarcastic sentences automatically from written text only, and from the reader's point of view. Managing this task without any knowledge of relevant contextual features, like prosody, is very hard.

The problem of sarcasm detection has been tackled with machine learning approaches, made possible by the availability of several annotated corpora. In the literature we can find two main categories of such corpora: \emph{automatically} annotated and \emph{manually} annotated.\\
The automatically annotated corpora are usually collected from the microblogging platform Twitter \cite{gonzalez2011identifying,reyes2013multidimensional} by exploiting the final hashtag of tweets. 
For instance, a tweet is labeled as sarcastic only if it ends with a hashtag such as \#sarcasm or \#irony. The same cue is used in Davidov, Tsur and Rappoport \shortcite{davidov2010semi} to produce a silver standard for evaluating their model. \\
Manually annotated corpora are collected from a more diversified range of social media, such as Amazon reviews \cite{filatova2012irony,reyes2011}, Reddit (Wallace et al. 2014) or online forums \cite{oraby2016creating,walker2012corpus}, and then labeled by hiring people in the Amazon Mechanical Turk portal. 
When using crowdsourcing, the annotation procedures are complex and involve, among others, a stage for ensuring that the workers understood the task and they are performing correctly, and a quality assurance stage for removing texts for which a high discrepancy between the annotators arises.

In this work we have tackled the problem of sarcasm detection by trying to use an entirely data-driven approach, exploiting a distributional semantics representation by inducing a semantic space and then applying a set of classifiers to classify the texts as being sarcastic or not sarcastic. With ``fully data-driven'' we mean approaches that are capable of finding connections between input text and class labels without using any a priori knowledge about the features that characterize a sarcastic statement.\\
%
In particular, we do not define ``irony'' or ``sarcasm'', neither use any definition. We simply 
rely on sets of sentences binary labeled for sarcasm detection
taking for granted that the labels correctly identify a sarcastic sentence.\\
It is worthwhile to point out that in this work we do not create any dataset: we simply exploit the labels of datasets that have already been produced by others, trying to give a baseline for the sarcasm detection task.

The contribution of this work can be summed up in three key points:
\begin{itemize}
\item[1)] we show that several machine learning methods can produce satisfactory results on automatic sarcasm detection by using distributional semantics to address the problem with no feature engineering;
\item[2)] we establish baselines for the sarcasm detection problem that could serve to the community to test newly proposed methods by releasing the code that we used to deal with the datasets; \item[3)] we establish the similarity of sarcastic content between pairs of corpora. 
\end{itemize}
To reach these goals, we exploit a Distributional Semantics approach, whose aim is to give a representation of words in a continuous vector space \cite{deerwester1990indexing,mikolov2013distributed}, where word similarity is coded in an unsupervised manner. This representation is useful for building models with little, or no, a-priori knowledge about the task \cite{collobert2011natural}. \\
Distributional semantics is a research field that concerns methodologies aimed at determining semantic similarities between linguistic items. 
The key idea is based on the hypothesis that words co-occurring in similar contexts tend to have similar meaning \cite{Harris1954,turney2010}.
Distributional semantics deals with the automatic construction of semantic
models induced from large unstructured textual corpora, and it exploits vector space models to represent the meaning of a word \cite{sales2016sem}.
Many methods can be applied to construct distributional
models. They range from the statistical models to machine learning ones \cite{dumais1988,mikolov2013distributed,pennington2014,astudillo2015}.
Among these techniques, Latent Semantic Analysis (LSA) is a methodology for building distributional semantic spaces that extract statistical relations between words which co-occurr in a given context though the use of the Truncated Singular value decomposition (T-SVD). 
In this work we explored and studied the possibility of building a data-driven model in the field of sarcasm detection exploiting the well-known Latent Semantic Analysis (LSA) paradigm both in its traditional formulation given by Landauer, Foltz and Laham \shortcite{landauer1998introduction} and  by using the Truncated Singular Value Decomposition (T-SVD) as a statistical estimator as illustrated in Pilato and Vassallo \shortcite{pilato2015tsvd}. \\
Both approaches have been used to create data-driven semantic spaces where documents and, generally, text chunks can be mapped. \\
The theory behind LSA states that the ``psychological similarity between any two words is reflected in the way they co-occur in small sub-samples of language'' (Landauer et al. 1998). \\
We have chosen to exploit the LSA paradigm since it is a well-known distributional semantics paradigm capable of modeling many human cognitive abilities; furthermore, it has many potential practical applications \cite{bellegardaB,deerwester1990indexing,foltz92,landauer99plato}. Moreover, it has been demonstrated in Pilato and Vassallo \shortcite{pilato2015tsvd} that Truncated Singular Value Decomposition (T-SVD), as used in LSA, can be interpreted as a statistical estimator, giving a robust theoretical interpretation to the Latent Semantic Analysis paradigm. Many researchers have successfully applied this technique for typical Semantic Computing applications, such as natural language understanding, cognitive modeling, speech recognition, smart indexing, anti-spam filters, dialogue systems, and other Statistical Natural Language processing problems \cite{bellegardaA,koeman14,pilato2015tsvd}. Moreover, Latent Semantic Analysis has been successfully used for inducing data-driven ``conceptual'' spaces \cite{Vassallo10}. 
Besides, it has been shown in \cite{altszyler16} that while a word embedding technique such as Word2vec outperforms LSA in medium-sized corpora, its performance considerably decreases when the corpus size is smaller; in that case, LSA gives better results.
For the aforementioned reasons, we have chosen this approach as a baseline for the detection of sarcasm in texts.
Furthermore, our study makes use of four machine learning methods that have been used on four manually annotated, publicly available corpora.\\
The experimental results show that our data-driven approach consisting of LSA followed by a classifier can establish models that outperform the published results on two of the corpora; additionally, it produces competitive results for the other corpora that we used for our evaluation.

The next section describes the state of the art in the field, Section \ref{methods} describes the Semantic Representation and the Machine Learning methods used in the study. Section \ref{datasets} introduces the datasets used for the experiments. Section \ref{experiments} summarizes the experimental results, Section \ref{conclusion} is for the final conclusions and remarks.

The code and the datasets used for the experiments are available on github\footnote{https://github.com/mattiadg/Sarcasm-LSA}.

\section{Related works}
The problem of sarcasm detection has been tackled using a wide range of supervised or semi-supervised techniques applied to corpora from different social media sources.

In the present work, we do not collect a new corpus for sarcasm detection, but sarcastic corpus annotation has received much attention in the literature. Most of the works have used unsupervised or semi-supervised approaches in order to reduce the cost of the annotation, while partially sacrificing the data quality. 
One of the first approaches was introduced by Tsur, Davidov and Rappoport \shortcite{tsur2010icwsm} for a corpus extracted from Twitter and further developed in Davidov et al. \shortcite{davidov2010semi} with a corpus consisting of Amazon reviews. This semi-supervised approach uses ``YAHOO! BOSS'' API web search for collecting $66,000$ utterances similar to the ones in a small initial labeled seed set. It was the first work to show that automatically-crawled data are useful for the task of sarcasm detection.
Most of the works have been pursued using data extracted from Twitter, as it is relatively easy to extract ironic or sarcastic tweets using the search by hashtag. 
In fact, in Twitter, the restricted number of characters allowed encourages to mark the ironic intent with a hashtag like \#irony or \#sarcasm to prevent ambiguities. 
The hashtag is usually removed from the tweets and used as a label for the silver standard.
Moreover, the first studies on Twitter data showed that the task is quite difficult also for human beings.
Gonz{\'a}lez-Ib{\'a}nez et al. \shortcite{gonzalez2011identifying} collected a corpus of $2,700$ tweets balanced between sarcastic, positive sentiment and negative sentiment. They presented a part of the corpus to human judges, who achieved low agreement and low accuracy.
Reyes et al. \shortcite{reyes2013multidimensional} collected a corpus using $4$ hashtags that identify four different categories, irony, education, humor, and politics, with $10,000$ tweets each. The same corpus was used in a later work~\cite{reyes2014difficulty}. Their results suggest that detecting sarcasm in full documents is easier than in single sentences because of the presence of a context, but in both cases, it remains a difficult task also for humans who often have a low agreement.
The specific case of positive sentiment and a negative situation, which is the most typical sarcastic situation, has also been analyzed \cite{riloff2013sarcasm}. In particular, the authors  found that less than half of the tweets ending with the hashtag \#sarcastic are recognized as sarcastic by humans after removing the hashtag.
Bharti, Babu, and Jena \shortcite{bharti2015parsing} proposed two algorithms with the goal to find, respectively, tweets with contrast in sentiment and situation, and tweets starting with interjections. They also found that the label distribution does not correlate perfectly with the hashtag distribution, e.g., only $1,200$ out of $1,500$ tweets ending with \#sarcastic are considered sarcastic by the annotators.
Farias, Patti and Rosso \shortcite{farias16} proposed a method that uses affective content to classify sarcastic tweets, and show that it outperforms preceding methods in several Twitter benchmarks.
%
Since classifying tweets by using only the text is a difficult task also for humans, other works proposed new methods capable of exploiting other kind of data, like the identity of the author or the thread of the tweet. 
Bamman and Smith \shortcite{bamman2015contextualized} augmented the feature vectors with features describing the author of the tweet and the user to which the tweet is addressed, obtaining significant improvements in accuracy. They also found that the hashtags \#sarcasm and \#sarcastic are mainly used when the audience is not known.
Wang, Wu, Wang and Ren \shortcite{wang2015twitter} use a sequential classifier for classifying tweets taking into account the previous responses, thus improving the performance concerning a simple multi-class classifier. \\ 
Amir, Wallace, Lyu, Carvalho and Silva \shortcite{amir2016modelling} used the dataset collected in Bamman et al. \shortcite{bamman2015contextualized} (which was not completely available) for training a \textit{deep learning} model that could represent users with \textit{user embeddings} and this method seems to outperform the method from Bamman and colleagues. 
Sarcasm classification on Twitter involves different modelling techniques that perform better when taking into account the user and the thread history of a Tweet. Our work focuses on the task of classifying a single document written by a single author. Thus, we focus mainly on different kinds of datasets.
Buschmeier, Cimiano and Klinger \shortcite{buschmeier2014impact} have studied the corpus introduced in Filatova \shortcite{filatova2012irony} by extracting a high number of features about typographic cues that can represent sarcasm, and used different classification methods obtaining results that vary significantly according to the classifier. They found that the single most important feature is the star rating of the review, and this happens because sarcastic reviews are more probable when a user did not like the product.\\
Wallace et al. \shortcite{wallace2014humans} created a corpus from Reddit posts, for which they also stored context information, such as the post that is answered. The authors proposed a method that uses the bag of words and other features from previous studies for building an SVM classifier that gets very low results.
Moreover, a correlation is found between posts for which the humans require the context and sarcastic posts. 
This can be explained by considering that the chosen sub-reddits\footnote{thematic forums in the Reddit platform} are about religion or politics, and they are thus very prone to controversial discussions. Consequently, to understand the ironic intent of a post it is quite important to know the author position on the topic and also the posts they are answering to. \\
Joshi, Sharma and Bhattacharyya \shortcite{joshi-sharma-bhattacharyya:2015:ACL-IJCNLP} used features for capturing intrinsic and extrinsic incongruity in texts and outperforms two previous methods both in tweets and in forum posts.
These works represent a valuable means of comparison for the present work. We show that an approach based only on distributional semantics is competitive with other approaches using more elaborated feature engineering, even when the data amount is quite small. 
Distributional semantics became popular in NLP thanks to the availability of good quality word embeddings~\cite{mikolov2013distributed}, and are introduced by design in deep learning models. In sarcasm detection, distributional semantics has been used to serve different roles. Ghosh, Guo, and Muresan \shortcite{ghosh2015sarcastic} have adopted word embeddings to disambiguate a literal use of single words from a sarcastic use.
Joshi, Tripathi, Patel, Bhattacharyya and Carman \shortcite{joshi2016word} use word embeddings to compute incongruities among words using them as additional features for methods selected from the literature. 
Our work differs from these as we use LSA instead of word embeddings, and distributional semantics is the only kind of features we use. 
Ghosh and Veale~\shortcite{ghosh2016} use LSA to extend the list of hashtags to find more sarcastic tweets on Twitter and use a deep neural network to perform the actual classification. Our work differs from theirs as we use LSA to compute the vectorial representation of documents and we do not perform tweet crawling. 
Poria, Cambria, Hazarika and Vij~\shortcite{cambria2016} train a convolutional neural network to classify sarcasm in tweets. They extend the neural network with features extracted from other datasets for sentiment, emotion and personality classification, as these features are considered to be useful for the task of sarcasm detection. 

\section{Data-Driven Induction of Semantic Spaces and Traditional Classifiers}
\label{methods}
We focused our research on the role that fully data-driven models can play in detecting sarcasm.
 To reach this goal, we exploited the Latent Semantic Analysis paradigm both in its traditional formulation (Landauer et al. 1998) and by using the Truncated Singular Value Decomposition (T-SVD) as a statistical estimator as shown in Pilato et al. \shortcite{pilato2015tsvd}. 
We have chosen to use the LSA paradigm to exploit a well-known and well-founded approach for inducing semantic spaces that have been effectively used in natural language understanding, cognitive modeling, speech recognition, smart indexing, and other statistical natural language processing problems.
The sub-symbolic codings of documents obtained by the aforementioned LSA-based approaches are then used as inputs by a set of classifiers to evaluate the differences of performances obtained by using different machine learning approaches and testing them on different sarcasm-detection datasets. 

The full work-flow, composed of the following steps:
\begin{itemize}
	\item Preprocessing of text (see Section \ref{preprocessing} for details).
	\item Data driven induction of semantic spaces by means of LSA-oriented paradigms (see Section \ref{datadriven}).
	\item Mapping new documents to the semantic space (see Section \ref{mapping}).
	\item Supervised Learning (see Section \ref{learning}).
\end{itemize}
does not require any expert or domain knowledge.

\subsection{Preprocessing of text}
\label{preprocessing}
The first step of preprocessing for texts is the tokenization using spaces, punctuation and special characters (e.g., \$, \euro, @) as separators. Thus one token is a sequence of alphanumeric characters or of punctuation symbols.
The set of all the extracted tokens constitutes a ``vocabulary'' named $V$.\\
The sequences of tokens, each representing a single document in the training set, are used to generate a word-document co-occurrence raw matrix $\mathbf{A}$, where each $(i,j)$ cell contains the number of times the token $i$ appears in the document $j$. Let $m$ be the number of tokens, i.e., $dim(V)=m$, and let $n$ be the number of documents of the corpus used for computing the matrix $\mathbf{A}$; the dimensionality of $\mathbf{A}$ is $m \times n$.  
\subsection{Data driven induction of semantic spaces by means of LSA-oriented paradigms}
\label{datadriven}
The matrix $\mathbf{A}$ is used and further processed to induce proper Semantic Spaces where terms and documents can be mapped. To generate these semantic spaces, we have used both the traditional LSA algorithm (Deerwester et al. 1990, Landauer et al. 1998) and the approach which uses T-SVD as a statistical estimator as proposed in Pilato et al. \shortcite{pilato2015tsvd}. For the sake of brevity, we call this last approach \emph{Statistical LSA} to differentiate it by the \emph{Traditional LSA}.
It is worthwhile to point out that, in the Latent Semantic Analysis paradigm (i.e., both ``general'' and ``statistical''), the corpus used for building the semantic space plays a key role in performances. As a matter of fact, large and heterogeneous corpora may give more noise or too much specific information from a single domain, decreasing the accuracy of the induced models \cite{Crossley2017}.

\subsubsection{Traditional LSA}
The traditional LSA is a procedure that has been used mainly for information retrieval (Deerwester et al. 1990). The previously described matrix $\mathbf{A}$ is used for computing a Tf-Idf (Term-Frequency Inverse-document frequency) matrix $\mathbf{M}$ \cite{sparck1972statistical}.
Let $k$ be the rank of $\mathbf{M}$. The following factorization, called \emph{Singular Value Decomposition} (SVD) holds for the matrix $\mathbf{M}$:

\begin{equation}
\mathbf{M}=\mathbf{U} \mathbf{\Sigma} \mathbf{V}^T
\end{equation}
where $\mathbf{U}$ is a $m \times k$  orthogonal matrix, $\mathbf{V}$ is a   $n \times k$  orthogonal matrix and $\mathbf{\Sigma}$   is a  $k \times k$  diagonal matrix, whose diagonal elements $\sigma_1,\sigma_2,\cdots ,\sigma_k$   are called \emph{singular values} of $\mathbf{M}$.  
It can be shown that the singular value decomposition of $\mathbf{M}$ is unique up to the order of the singular values and of the corresponding columns of $\mathbf{U}$ and $\mathbf{V}$, so there is no loss of generality if we suppose that $\sigma_1,\sigma_2,\cdots ,\sigma_k$  are ranked in decreasing order. \\
Let $r$ be an integer such that $r < k$, let  $\mathbf{U}_r$  be the matrix obtained from $\mathbf{U}$ by removing its last $k-r$   columns,  $\mathbf{V}_r$  the matrix obtained from $\mathbf{V}$ in the same manner and  $\mathbf{\Sigma}_r$ the diagonal matrix obtained from  $\mathbf{\Sigma}$  by suppressing both its last $k-r$   rows and $k-r$  columns. $\mathbf{U}_r$ is the matrix containing the $r$-dimensional vector representation of the words and $\mathbf{V}_r$ is the matrix containing the $r$-dimensional vector representation of the documents. It can be shown (Deerwester et al. 1990) that the matrix: 

\begin{equation}
\mathbf{M}_r=\mathbf{U}_r \mathbf{\Sigma}_r \mathbf{V}_r^T
\end{equation}
is the best rank $r$ approximation to $\mathbf{M}$ according to the Frobenius distance\footnote{Given two $M \times N$ matrices $\mathbf{A}=[a_{ij}]$ and $\mathbf{B}=[b_{ij}]$, their Frobenius distance is defined by: \[ d_F(\mathbf{A},\mathbf{B})=\sqrt{\sum_{i=1}^M \sum_{j=1}^N   ( b_{ij}-a_{ij} )^2} \] }.   
$\mathbf{M}_r$ is called the \emph{reconstructed matrix}. The process by which $\mathbf{M}_r$   is obtained from $\mathbf{M}$  is called \emph{Truncated Singular Value Decomposition} (T-SVD). The book by Golub and Van Loan \shortcite{golub1996matrix} provide further details about the Singular Value Decomposition technique.

\subsubsection{Statistical LSA}
The traditional Latent Semantic Analysis based on T-SVD is one of the possible methods to infer data-driven models. Furthermore, one of its major drawbacks, which is the lack of a sound statistical interpretation, has been recently overcome in Pilato et al. \shortcite{pilato2015tsvd}, where authors have been presented a statistical explanation of this paradigm. 

According to this interpretation, the T-SVD algorithm, as used in the Latent Semantic Analysis paradigm, acts as an \textit{estimator}, which conveys statistically significant information from the sample to the model.

To briefly sum-up the procedure, we recall here the concepts of \emph{probability amplitude} and \emph{probability distribution} associated with a matrix as they have been defined in Pilato et al. \shortcite{pilato2015tsvd}.

Let $M$, $N$ be two positive integers and let $\mathbb{R}$ be the set of real numbers.
Given a $M \times N$  matrix $\mathbf{B}=[b_{ij}]$ with $b_{ij} \in \mathbb{R}$, $i\in [1,2,\cdots,M]$, $j\in [1,2,\cdots,N]$ where at least one of its components $[b_{ij}]$ is positive, 
we define a set $J$, composed of all the pairs $(i,j)$ that identify the positive components of $\mathbf{B}$, i.e.:
\begin{eqnarray}
J=\{ (i,j):b_{ij}> 0 \} ~~i\in [1,2,\cdots,M], j\in [1,2,\cdots,N]
\end{eqnarray}    
Subsequently, we define the \emph{probability amplitude} associated with $\mathbf{B}$, the $M \times N$ matrix $\mathbf{\Psi} =[\psi_{ij}]$ resulting from the mapping $p_a(\cdot)$:

\begin{equation}
\mathbf{\Psi} \equiv p_a(\mathbf{B}): \mathbb{R}^{M \times N} \rightarrow [0,1]^{M \times N}
\end{equation}
whose elements $[\psi_{ij}]$ are computed as:

\begin{eqnarray}
\psi_{ij} = \left\{
\begin{array}{lr}
\frac{b_{ij}}{\sqrt{\sum_{(i,j)\in J}b_{ij}^2}} & if~~ b_{ij} > 0\\
\\
0 & if~~ b_{ij} \le 0
\end{array}
\right.
\end{eqnarray}        
so that $\forall(i,j)$ it is $\psi_{ij}\ge 0$ and $\sum_{i=1}^{M}\sum_{j=1}^{N}\psi_{ij}^2=1$.

We define also the \emph{probability distribution} associated with a matrix $\mathbf{B}$ the $M \times N$ matrix resulting from the mapping $p_d(\cdot )$:

\begin{equation}
\mathbf{B}^{(2)} \equiv p_d(\mathbf{B}): \mathbb{R}^{M \times N} \rightarrow \mathbb{R}^{M \times N}
\end{equation}
whose elements are the squares of the elements of $\mathbf{B}$, i.e. $\mathbf{B}^{(2)}=[b_{ij}^2]$.
The method starts with a raw data matrix $\mathbf{A}$ consisting of positive values. In our study the raw data matrix $\mathbf{A}$ is the term-document co-occurrence matrix. From $\mathbf{A}$ a real-valued normalized matrix $\mathbf{Q}$ is computed by dividing every element for the sum of all elements of $\mathbf{A}$. 
\begin{equation}
\mathbf{Q} = \frac{\mathbf{A}}{\sum\sum a_{ij}}, \forall a_{ij} \epsilon \mathbf{A}.
\end{equation}

If we call $\mathbf{\Psi}_Q$ the matrix:

\begin{equation}
\mathbf{\Psi}_{Q} = \Big[ \sqrt{q_{ij}} \Big]
\end{equation}

The matrix $\mathbf{\Psi}_Q$ can be decomposed with the SVD technique:
\begin{equation}
\mathbf{\Psi}_Q=\mathbf{U}\mathbf{\Sigma}\mathbf{V}^T
\end{equation}
and its best \emph{rank-r} decomposition $\mathbf{\Xi}=[\xi_{ij}]$ is obtained by applying the T-SVD technique, which minimizes the Frobenius distance $d_F(\mathbf{\Xi},\mathbf{\Psi}_Q)$, given $r$:

\begin{equation}
\mathbf{\Xi}=\mathbf{U}_r\mathbf{\Sigma}_r\mathbf{V}_r^T
\end{equation}

Even if $\mathbf{\Xi}$  is not a probability distribution, the computation of $\mathbf{\Xi}$ makes it possible to identify, without any further addition of external information, the probability distribution we are looking for. As shown in Pilato et al. \shortcite{pilato2015tsvd}, it theoretically suffices computing the probability amplitude associated to $\mathbf{\Xi}$, i.e. $p_a(\mathbf{\Xi} )$, and consequently calculating the probability distribution $p_d(p_a(\mathbf{\Xi}) )$ associated to  $p_a(\mathbf{\Xi} )$. The aforementioned Frobenius distance $d_F(\mathbf{\Xi},\mathbf{\Psi}_Q)$ constitutes an upper bound to the Hellinger distance\footnote{Given two  $M \times N$ matrices $\mathbf{A}=[a_{ij}]$ and $\mathbf{B}=[b_{ij}]$, their Hellinger distance is defined by:\[ d_H(\mathbf{A},\mathbf{B})=\sqrt{\sum_{i=1}^M \sum_{j=1}^N   ( \sqrt{b_{ij}}-\sqrt{a_{ij}} )^2} \]
} between the sample probability $\mathbf{Q}$ and the probability distribution estimated by the procedure.

\subsection{Mapping new documents to the semantic space}
\label{mapping}
Both LSA approaches illustrated in the previous subsections provide us with the three, obviously different for each approach, matrices $\mathbf{U}_r$, $\mathbf{\Sigma}_r$ and $\mathbf{V}_r$.\\
The $\mathbf{\Sigma}_r$ and the $\mathbf{U}_r$ matrices can be used for computing the vector representation of the new documents into the induced semantic space. 
The $\mathbf{\Sigma}_r$ matrix contains in its diagonal the singular values; $\mathbf{U}_r$ is composed by rows that represent the r-dimensional sub-symbolic, i.e., numerical, mapping in the semantic space of the tokens constituting the vocabulary $V$. 
Then, given a text chunk $d$, $d$ is sub-symbolically  represented by a $dim(V)$-dimensional word occurrence vector $\mathbf{d}$, from which it is computed a vector $\mathbf{q}$ with two different procedures depending on which LSA paradigm has been chosen.\\
In the case of Traditional LSA, it is the Tf-Idf representation \cite{salton1988term} of $\mathbf{d}$ by using the same parameters learned during training. \\
In the case of the Statistical LSA, the $\mathbf{d}$ vector is transformed into $\mathbf{q}$ similarly as the matrix $\mathbf{A}$ is transformed into the matrix $\mathbf{Q}$:
\begin{equation}
\mathbf{q}=\sqrt{\frac{\mathbf d}{\sum_i\sum_j a_{ij}}} ~~~~ \forall i\in  \left[1,dim(V)\right]
\end{equation}
Once the appropriate coding of $\mathbf{q}$ has been computed, an r-dimensional vector $\mathbf{d}_r$ representing the sub-symbolic coding of $d$ is then obtained from the vector $\mathbf{q}$ by means of the following mapping formula:
\begin{equation}
\mathbf{d}_r  = \mathbf{q}^T \mathbf{U}_r \mathbf{\Sigma}_r^{-1} 
\end{equation}

\subsection{Supervised learning}
\label{learning}
The training and test documents are mapped into the semantic spaces induced at the previous step. These vectors, sub-symbolic coding of the documents, are therefore used as inputs to different classifiers to train or test on them. Such classifiers will finally solve a binary classification problem assigning the label $1$ (sarcastic) or $0$ (nonsarcastic) to a generic document.
For this study we have used Support Vector Machines, Logistic Regression, Random Forests, and Gradient boosting as they represent the state of the art for most of the binary classification problems with small datasets. In the following, we recall a brief description of them. 
\subsubsection{Logistic Regression}

The \emph{logistic regressor} (LR) is a generalized linear
model suitable for binary responses \cite{GLMbook1989}. In LR
the following log-linear model is adopted:

\begin{equation}
ln \left({\frac{p}{1-p}} \right)= \boldsymbol\beta^t\mathbf{x} 
\end{equation}

where $p$ represents the probability of the success outcome. A suitable way of minimizing the so called \emph{empirical risk} is the numerical estimation of the  $\mathbf{\beta}$s coefficient by a maximum likelihood procedure: 

\begin{equation}
E(\boldsymbol \beta) = \sum_{i=1}^m [-y_i \ln p + (1-y_i) \ln (1-p)] + \lambda \vert\vert \boldsymbol\beta \vert\vert 
\label{minimization}
\end{equation}

where $(\mathbf{x_i},y_i)$ is the training set, $\vert\vert \mathbf{\beta}\vert\vert$ is the norm of the weights vector used for regularization, and can be either the $L1$ or the $L2$ norm, and $\lambda$ is the weight to give to the regularization factor.
The function in formula \ref{minimization} is convex, so it can be minimized even with the simple gradient descent algorithm, but more complex algorithms can be used in order to reduce the convergence time. In this work we use the trust region Newton method proposed by Lin, Weng and Keerthy~\shortcite{lin2008trust}, as provided by the LIBLINEAR library~\cite{fan2008liblinear}.

\subsubsection{Support Vector Machine}
A \textit{kernel} $K$ is any mapping satisfying 
\begin{equation} K(\mathbf{x},\mathbf{y})= < \phi(\mathbf{x}), \phi(\mathbf{y}) > 
\end{equation}
where $\mathbf{x}$,$\mathbf{y}$ are elements in the input space, $\phi$ is a mapping from the input space to a new representation space $F$ where an inner product is defined. 
The function $\phi$ is chosen to be nonlinear, and the dimension of the feature space is taken intentionally greater than the dimension of the input space.  
These choices could give the chance to make the classification problem linearly separable in $F$.
Support vector machines (SVMs), also called \emph{kernel machines} \cite{cristianini2000introduction} are binary
linear classifiers that make use of kernels. They search for the optimal hyperplane $\hat{h}$
in the feature space that maximizes the geometric margin, which
is the distance of the hyperplane to the nearest training
data point of any class. The main advantage
of SVM is that it provides a solution to the global
optimization problem, thereby reducing the generalization
error of the classifier.
The formulation of SVM can be easily extended
to build a nonlinear classifier by incorporating a
\emph{kernel} of the class H 

\begin{equation} H=\{sgn(K(\boldsymbol\beta,\mathbf{x})):\mathbf{\beta} \in \mathbb{R}^n \} 
\end{equation}

No systematic tools have been developed
to automatically identify the optimal kernel for a
particular application.

\subsubsection{Random Forest}
Decision trees \cite{kotsiantis13} are rooted trees that can be 
used successfully as classifiers \cite{cartbook}.
Each node of the three represents a binary rule that splits the
feature space according to the value of a predictive
feature and a path from the root to leaf nodes represents
a series of rules that are used to recursively divide
the feature space into smaller subspaces, where a
class label is assigned. The structure of the tree in terms of split nodes can be learned from data by using several approaches. 
\emph{Random forests} \cite{breiman11} are an ensemble of decision trees, found using the bootstrap sampling technique on the training set.
In particular, a fixed number of random samples are extracted with replacement from the training set, and each of them is used as a training set to fit a decision tree. The forest is composed by each of these decision trees, and the final predictions are made by averaging the predictions from all the individual decision trees.   

\subsubsection{Gradient boosting}
Boosting is another ensemble strategy with the special purpose of improving the combination of a set of weak classifiers.
These are chosen to be of very low model complexity such as the case of decision trees with a single split. The general framework of boosting sequentially adds a tree to an ensemble, the new one with the goal of correcting its predecessor.
\emph{Gradient boosting} \cite{xgb16} uses a gradient-descent like procedure to sequentially improve a tree classifier. This is done by adding to the actual classifier a new decision tree learned from the residual errors made by the predecessor. The final predictions are made by the tree classifier resulting after a fixed number of iterations of the procedure.  

\section{Datasets}
\label{datasets}
We have chosen $4$ corpora for our experiments, all of them are publicly available and treating the problem as a binary classification: ``SarcasmCorpus''\footnote{http://storm.cis.fordham.edu/filatova/SarcasmCorpus.html} (Filatova 2012) , ``IAC-Sarcastic''\footnote{https://nlds.soe.ucsc.edu/node/32} \cite{lukin2013really}, which is a subset of Internet Argument Corpus1.0 prepared for sarcasm detection, ``irony-context''\footnote{https://github.com/bwallace/ACL-2014-irony} (Wallace et al. 2014), and ``IAC-Sarcastic-v2'' (Oraby et al. 2016), which is extracted from the second version of Internet Argument Corpus\footnote{https://nlds.soe.ucsc.edu/iac2}\cite{abbott2016internet}. 
In order to provide a more complete evaluation, we also use the corpus of the shared task ``Semeval2018 Task 3A''\cite{semeval2018}.
\subsection{SarcasmCorpus}
Filatova \shortcite{filatova2012irony} collected 1254 reviews from Amazon for different kinds of products, of which 437 are sarcastic, and 817 are not sarcastic. 
Each review in the corpus consists of the title, author, product name, review text and number of stars, and the review is a stand-alone document referring to a single product.
This corpus, like all the others considered in this work, has been entirely hand-labeled by the Amazon Mechanical Turkers, who were asked whether each review contains sarcasm in it. Each text has been presented to 5 Turkers and has been classified as sarcastic when at least three among five workers agreed.
The corpus contains $21,744$ distinct tokens, with $5,336$ occurring only in sarcastic reviews, $9,468$ occurring only in literal reviews and $6,940$ occurring in both categories.
Buschmeier et al. \shortcite{buschmeier2014impact} made an interesting analysis of the corpus by collecting some statistics and publishing the only classification results that are available for it up to now. They extracted $29$ task-specific features and combined them with the bag-of-words representation and multiple classifiers. The bag of words was found to be important for the classification. In fact, for example, they get a poor 50.9\% F-score value with logistic regressor without bag-of-words, which is increased to 74\% by using it. This result is surely related to the difference in terms used by the two classes, but it also shows that information about the words used in the document is needed for the task.

\subsection{IAC-Sarcastic}
The second dataset we used is the IAC-Sarcastic sub-corpus, which consists of $1995$ posts coming from 4forums.com, a classical forum where several topics are discussed. This corpus is actually extracted from the larger Internet Argument Corpus (IAC), containing $11,800$ discussions, $390,704$ posts and $73,000,000$ words. In IAC there are $10,003$ Quote-Response (Q-R) pairs and $6,797$ three-posts chains that have been manually labeled for several HITs (Human-Intelligence Tasks) by Amazon Mechanical Turk. 
For each Q-R item, the Turkers were asked to evaluate the response section by considering the quote as a context. One of the HITs regarded the identification of a sarcastic response. As a result, the IAC-Sarcastic Corpus consists of 1995 responses, without any quote, with a binary label that indicates the presence of sarcasm.  998 texts are labeled as sarcastic, and 997 are not, so this is one of the rare balanced datasets for this task. To the best of our knowledge, only the work by Justo, Corcoran, Lukin, Walker, and Torres \shortcite{justo2014}  published results on the sarcastic task of the IAC dataset, but the authors made a different sampling of the documents from the one used for IAC-Sarcastic. Thus, our results for this corpus are not comparable with the ones reported in that work.

\subsection{Irony-context}
A third dataset is the one collected in Wallace et al. \shortcite{wallace2014humans}. The main goal of that study was to highlight the role of the context of a text to make irony understandable by humans.
The dataset is extracted from Reddit by collecting comments from the following six sub-reddits: politics, progressive, conservative, atheism, Christianity, technology, with their respective size of $873$, $573$, $543$, $442$, $312$ and $277$ samples.
Each comment has been labeled by three university undergraduates using a browser interface which let them see the context of the comment in the form of previous comments or related pages under request.
The label of a comment was selected with a simple majority of $2$ out of $3$ labelers. For each comment and each labeler, they stored whether the context has been requested and if the labeler changed his mind after having seen it. This allowed the authors to study the correlation between the sarcastic label and the requests for context. \\
The results allowed the authors to infer that the machines would also need the context for detecting sarcasm, as their model did not predict correctly the texts for which the humans required the context.
This is an important cue that should be considered while developing sarcasm detection methods, even though we do not explicitly consider the context of our method. As a result, we cannot expect to obtain high absolute results for this dataset by letting the model observe only the single text.

\subsection{IAC-Sarcastic-v2}
In 2016 a new version of IAC was made available (IACv2) (Abbot et al. 2016), and after some months also the sarcastic sub-corpus was released (Oraby et al. 2016),  which is bigger than the first version. It consists of three sub-corpora, among which the bigger one is called ``generic'', and it is made of $3,260$ posts per class collected from IACv2. For the creation of this sub-corpus, the authors produced a high-precision classifier for the non-sarcastic class, which helped to filter out many non-sarcastic posts from the original corpus and lower the labeling costs. Then, to have high-quality labeling, they required a majority of 6 out of 9 sarcastic annotations to label a post as sarcastic.\\
To produce a more diverse corpus, they built two more corpora focused on particular rhetorical figures often associated with sarcasm: rhetorical questions and hyperboles. For both of the sub-corpora, the authors used patterns to recognize posts containing the chosen rhetorical figure from IACv2. Each of the collected posts has been subsequently shown to five AMTs for the sarcastic/not sarcastic annotation. The label is given with simple majority.\\
The purpose of these two focused sub-corpora is to force classifiers to find some semantic cues which can distinguish sarcastic posts even in the presence of rhetorical figures usually associated with sarcasm. In fact, the presence of hyperboles has been used before as a feature for detecting sarcasm~\cite{buschmeier2014impact}.

\subsection{Semeval-2018 Task3 Corpus of Tweets}
\label{tweet-dataset}
The International Workshop on Semantic Evaluation Semeval-2018 featured a shared task on verbal irony detection in tweets (Van Hee et al. 2018). The corpus contains a class-balanced training set consisting of $3,833$ tweets, and a test set with $784$ tweets. In the test set, only 40\% of the instances are ironic. 
The corpus has been collected from Twitter searching for tweets with the hashtags \#irony, \#sarcasm and \#not.
The corpus has been annotated by three students in linguistics who showed a high inter-annotator agreement.
After the annotation, $2,396$ tweets out of $3,000$ were ironic and only $604$ were not. Thus, an additional set of $1,792$ non-ironic tweets was added to the corpus.
Finally, the corpus was split randomly in class-balanced training and test set, but an additional cleaning step for removing ambiguous sentences modified the proportion to 40\% ironic.

\section{Experiments and results}
\label{experiments}

\begin{table}
\caption{Results of the in-corpus experiments for two versions of SarcasmCorpus datasets. The difference is that SarcasmCorpus$^*$ use the additional information of the star rating. For each Method (in the first column) and each corpus, the result with highest average $F$- score on a $K$-fold cross-validation setting are reported. The related precision (prec) and recall (rec) scores are also reported. Rows $3$ and $4$ show the result of the two methods used for comparison. From row $5$, the results of the proposed models are reported.  In $\mathbf{bold}$ the best $F$-score, precision, recall for every considered corpus. The symbols S and T before each method indicate the adoption of Statistical or Traditional LSA respectively.}
	\begin{minipage}{\textwidth}
	\begin{tabular}{ccccccc}
	\hline\hline
		 & \multicolumn{3}{c}{SarcasmCorpus} & \multicolumn{3}{c}{SarcasmCorpus*}\\
	\hline
		Method &F    &prec          &rec   &F        &prec           &rec \\
	\hline
	 Buschmeier et al. \shortcite{buschmeier2014impact}  &68.8 &76.0 &63.3  &74.4     & 81.7  &68.9\\
     \hline
    Poira et al.~\shortcite{cambria2016}  & \textbf{73.7}  &  66.7  & \textbf{82.4}  &  -  &  -  &  -  \\
    \hline
    S-SVM gauss      &66.3          &70.8 &63.6            &73.3     &78.1           &69.4          \\
    S-Log.Reg.L1      &71.8 &69.5          & 74.6  &79.4            &77.5          &81.7\\
    S-RF    &  51.5 &  \textbf{81.2}  & 38.0  & 74.3  &  \textbf{82.2}  &  68.5\\
    S-XGB   &  64.5 & 72.9  & 59.1  &  77.3  &  80.3  &  74.6\\
	T-SVM gauss    &71.4          &70.9          &72.1            &75.7           &79.0      &73.0          \\
    T-Log.Reg.L1    &70.7          &70.4          &71.4            &\textbf{80.7}           &79.8      &\textbf{81.9}          \\
    T-RF    &  52.4  &  81.9  &  39.1  &  76.1  &  80.3  &  72.5 \\
    T-XGB   &  63.4  &  72.4  &  57.0  &  78.8  &  81.4  &  76.6\\
	\hline\hline
	\end{tabular}
	\vspace{-2\baselineskip}
	\end{minipage}
	\label{main-table1}
	\end{table}

{\small
\begin{table}
	\caption{Results of the in-corpus experiments on IAC-Sarcastic and Irony-Context. For each method (first column) and each corpus, the result with the highest average $F$-score on a $K$-fold cross-validation setting are reported. The related precision (prec) and recall (rec) scores are also reported. Rows $3$ and $4$ show the results of the two methods used for comparison. From row $5$, the results of the proposed models are reported. In $\mathbf{bold}$ the best $F$-score, precision, recall for every considered corpus. The symbols S and T before the methods indicate the adoption of Statistical or Traditional LSA respectively.}
	\begin{minipage}{\textwidth}
		\begin{tabular}{ccccccc}
			\hline\hline
		    & \multicolumn{3}{c}{IAC-Sarcastic} & \multicolumn{3}{c}{Irony-Context} \\
		    \hline
		     Method & F   & prec   & rec        & F        & prec      & rec\\
		    \hline
	Wallace et al. \shortcite{wallace2014humans}     &-             &-       &-  &38.3           & 31.5   & 49.6\\
\hline
Poira et al.~\shortcite{cambria2016}  &  \textbf{72.7} &  \textbf{75.3}  &  70.2  &  \textbf{58.1}  &  \textbf{48.5}  &  72.6\\
\hline
S-SVM gauss   &66.8 &57.6    & \textbf{79.5}      & 45.8         & 31.9      & \textbf{81.2}\\
S-Log.Reg.L1   &63.3          &61.6    & 65.2               & 46.0         & 35.2      & 66.5\\
S-RF  &  62.7 &  61.5  &  64.0  &  4.5  &  48.8  &  2.4\\
S-XGB &  61.2 &  59.3  &  63.5  &  23.7 & 43.6  &  16.4\\
T-SVM gauss &62.3          &64.6    & 60.2               & 42.8         & 36.7    & 51.6\\
T-Log.Reg.L1 &63.1          &64.5 & 61.9         & 45.9         & 38.5             & 56.7\\
T-RF   &  62.7  &  61.2  &  64.2  &  4.5  &  43.6  &  2.4 \\
T-XGB  &  64.2  &  62.3  &  66.5  &  19.9  &  37.4  &  13.8 \\
			\hline\hline
		\end{tabular}
		\vspace{-2\baselineskip}
	\end{minipage}
	\label{main-table2}
\end{table}
}

\begin{table}
	\begin{center}
	\caption{Results of the in-corpus experiments on IAC-v2. For each model (first column) and each corpus, the result with the highest average $F$- score on a $K$-fold cross-validation setting are reported. The related precision (prec) and recall (rec) scores are also reported. Rows $3$ and $4$ show the result of the two methods used for comparison. From row $5$, the results of the proposed models are reported. In $\mathbf{bold}$ the best $F$-score, precision, recall for every corpus. The symbols S and T before the methods indicate the adoption of Statistical or Traditional LSA respectively.}
	\begin{minipage}{\textwidth}
		\begin{tabular}{cccccccccc}
		\hline\hline
		& \multicolumn{3}{c}{IAC-v2 GEN} & \multicolumn{3}{c}{IAC-v2 HYP} & \multicolumn{3}{c}{IAC-v2 RQ} \\
		\hline
		Method & F   & prec   & rec & F   & prec   & rec        & F        & prec      & rec\\
		\hline
		Oraby \\
		et al.~\shortcite{oraby2016creating}
		  & 74.0 & 71.0 & 77.0 &70.0  &\textbf{71.0}        & 68.0  & 65.0    &68.0 &63.0\\
		\hline
        Poira et al.\\~\shortcite{cambria2016}  &  72.6  &  \textbf{77.5}  &  68.2  &  -  &  -  &  -  &  -  &  -  &  - \\
       \hline
		S-SVM gauss    &\textbf{74.9} &64.7        &\textbf{88.9} & 68.2                  & 66.1                 & \textbf{71.2}   & 69.2  & 67.0                 & 71.8\\
		S-Log. Reg.L1     & 74.6         & 72.4       & 77.0 & 68.0                  & 68.4                 & 67.7            & 69.8    & 69.8                 & 70.0\\
        S-RF  &  71.0  &  66.1  &  76.7  &  61.8  &  63.9  &  60.8  &  69.0  &  62.6  &  \textbf{77.2} \\
        S-XGB &  71.9  &  68.9  &  75.2  &  64.2  &  67.9  &  61.5  &  68.9  &  67.8  &  70.1 \\
		T-SVM gauss  & 71.9         & 73.9       & 71.9 & 67.7                  & 68.7                 & 67.0            & 70.9    & \textbf{76.6}                 & 66.4\\
		T-Log.Reg.L1  & 72.9         & 73.7                 & 71.9 & \textbf{72.4}                  & 69.5                 & 69.4            & \textbf{72.4}    & 72.4        & 72.7\\
        T-RF  &  70.6 &  70.1  & 71.3  &  60.7  &  61.1  &  60.9  &  71.5  &  69.1  &  74.4\\
        T-XGB &  72.2 &  71.2  &  73.3  &  63.7  &  60.5  &  67.7  &  70.5  &  69.6  &  71.5\\
		\hline\hline
		\end{tabular}
		\vspace{-2\baselineskip}
	\end{minipage}
	\label{main-table3}
	\end{center}
\end{table}

\subsection{Experimental setup}
\label{sec:exp_setup}
We ran three groups of experiments, to assess both the effectiveness of our approach when compared with the approaches we found in literature, and its capability of extracting features that are relevant for sarcasm in a cross-domain scenario. In both cases, we denote with the word \textit{model} one of the possible combinations of classic/statistical LSA and a classifier. The used classifiers are Support Vector Machine (SVM), Logistic regression (Log.Reg), Random Forest (RF) and gradient boosting (XGB).   

For the \textit{first} group of experiments, we evaluated the performance of each of our models in every corpus. We use $10$-fold cross-validation and report the mean values of $F$-score, precision, and recall among all the folds. The proportion of the two classes in each fold is equal to the proportion in the whole corpus. 
Where applicable, we compare our results with existing results in the literature. Besides, we compare with the method presented in Poira et al.~\shortcite{cambria2016}.\footnote{The experiments on all the considered datasets have been carried out by the authors of the paper, who have also provided the related results.}

The \textit{second} group of experiments has been performed on the Semeval 2018 Task 3 dataset (Van Hee et al. 2018). We first find the best LSA dimensionality by 10-fold cross-validation in the training set. Then, we trained again the models in the whole dataset and evaluated them in the test set for comparison with the participants to the shared task.

The \textit{third} group of experiments is inter-corpora. For each experiment, we have chosen one corpus as a training set and another one as a test set. This process is performed for all the models and all the corpora pairs. We aim to find whether sarcasm detection is domain-dependent. 

Finally, in the \textit{fourth} group of experiments (union experiments) we perform another 10-fold in which all the corpora are concatenated. Each fold contains samples from every corpus proportionally to the size of that corpus. The goal of this experiment is to understand whether simply adding more data, but from different domains, improves the classification performance.

The hyperparameters of the classifiers have been chosen by grid search on SarcasmCorpus with LSA dimensionality $40$, and then used for all the reported experiments.
We use SVM with Gaussian kernel, C value of $100$, $\sigma=1/r$ logistic regression with penalty L1 and C=$10$ and decision tree with entropy loss. SVM and logistic regression both have balanced class weights to cope with unbalanced datasets.

\begin{figure}[t]
	\centering
	\includegraphics[scale=0.53]{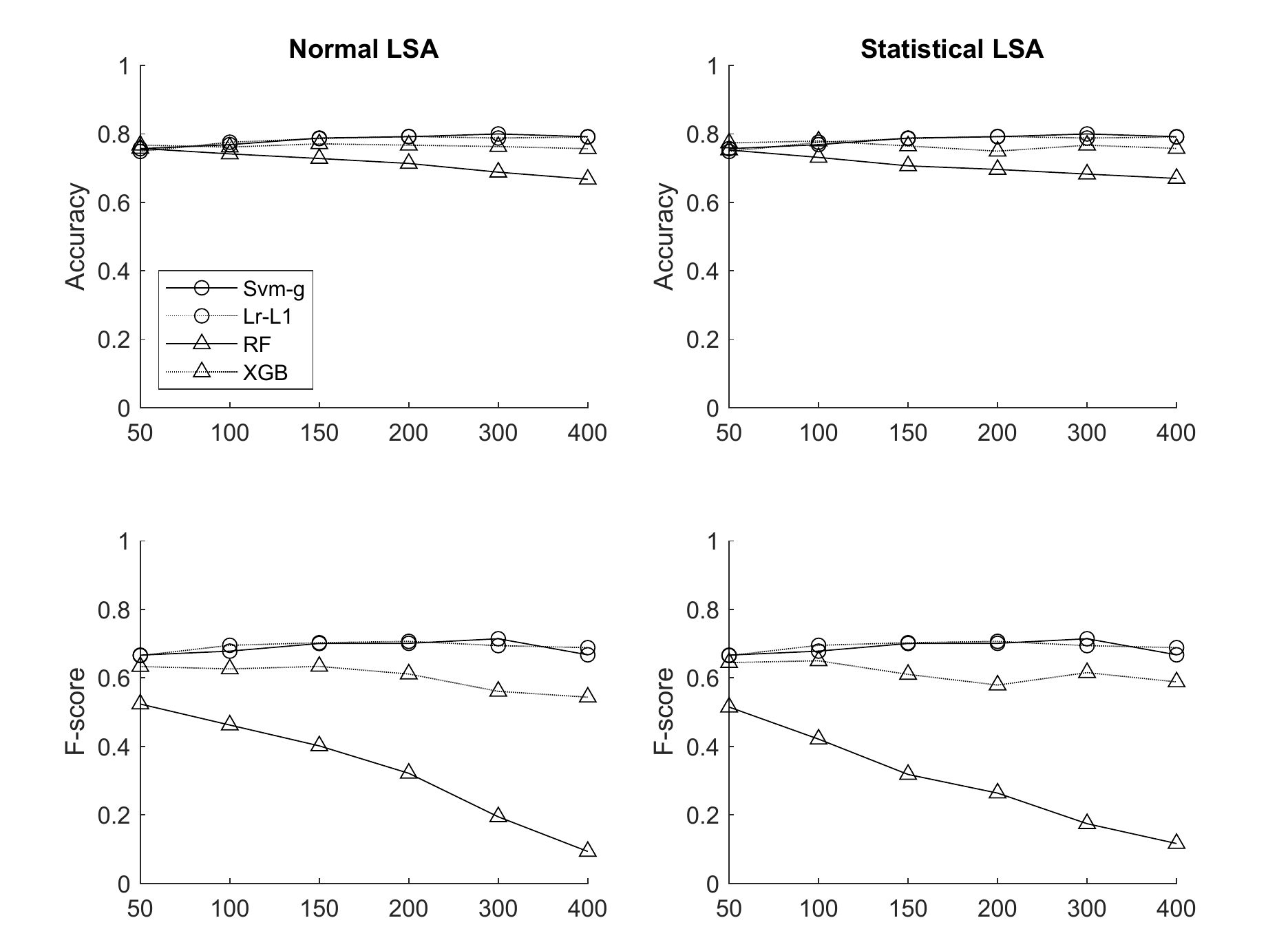}
	\caption{Accuracy and F-score values of all the considered classifiers for different LSA size in the case of Sarcasm Corpus}
	\label{fig:SarcasmCorpus}
\end{figure}

\begin{figure}[t]
	\centering
	\includegraphics[scale=0.53]{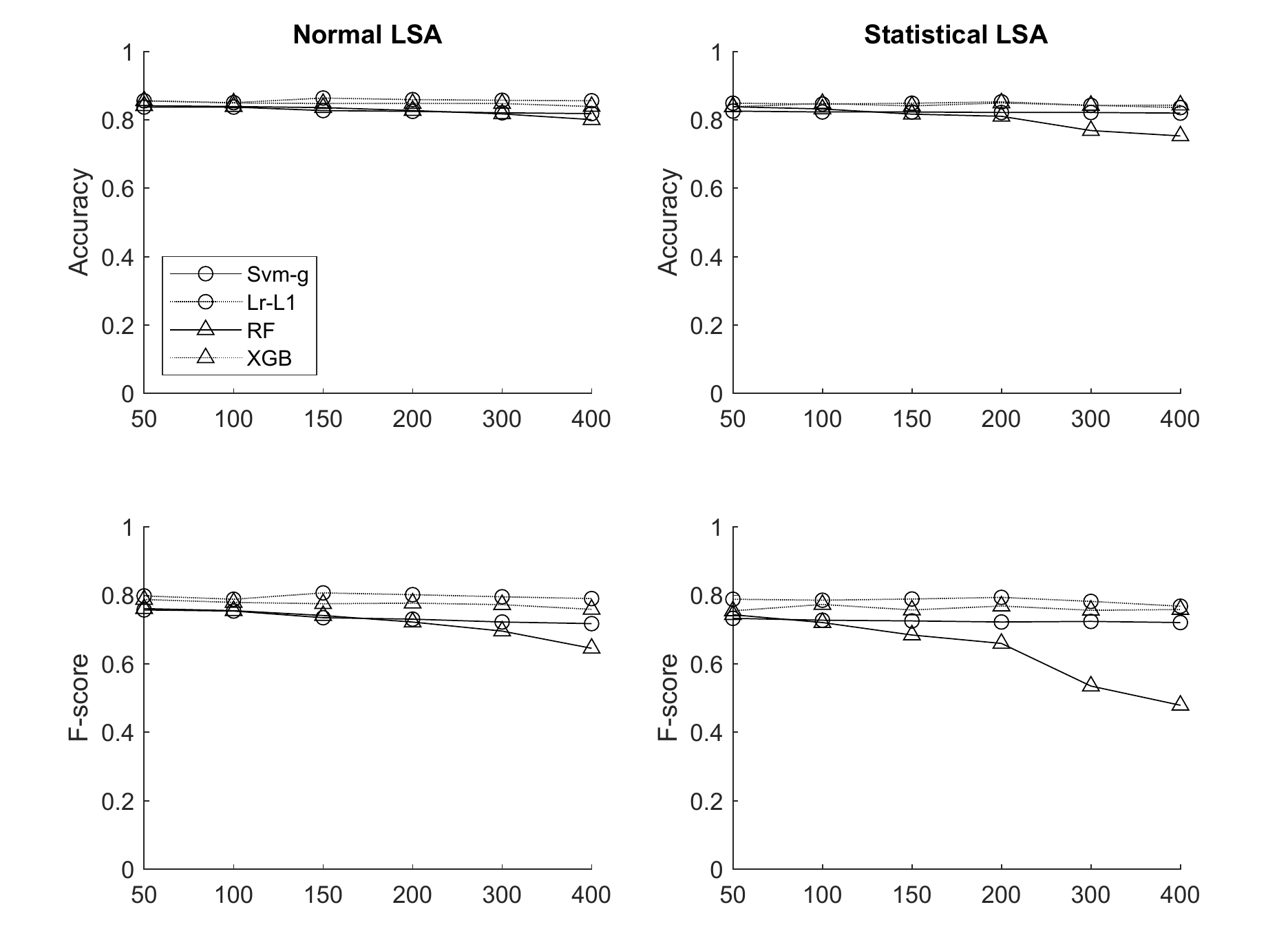}
	\caption{Accuracy and F-score values of all the considered classifiers for different LSA size in the case of Sarcasm Corpus*}
	\label{fig:SarcasmCorpusStar}
\end{figure}

\subsection{In-corpus Experiments}

\subsubsection{SarcasmCorpus}
In SarcasmCorpus each sample consists of a review title, a review text, a product name and the number of stars given to the product ranging from $1$ to $5$. 
Buschmeier et al. \shortcite{buschmeier2014impact} showed that the star rating is the most discriminative feature. Thus we performed the experiment both including and not including it. In Table~\ref{main-table1}, we refer to ``SarcasmCorpus'' when the star rating is not used, and ``SarcasmCorpus*'' when it is used.
We use the star rating by simply concatenating it to the document vector produced by LSA.
The document vector is computed only from the review texts because in our preliminary experiments we found that the other parts are not useful for the task.
Accuracy and F-score values of all classifiers for SarcasmCorpus and SarcasmCorpus* are plotted in Figures \ref{fig:SarcasmCorpus} and \ref{fig:SarcasmCorpusStar}, and the best F-scores, with the relative precision and recall, are reported in the two columns SarcasmCorpus and SarcasmCorpus* of Table \ref{main-table1}. 
The best result from the logistic regression in SarcasmCorpus is $71.8$ which represents a $4.4$\% relative improvement concerning the $68.8$ reported in the above-mentioned work by Buschmeier et al. \shortcite{buschmeier2014impact}. The results from Poira et al. \shortcite{cambria2016} are even higher in terms of F-score, with a relative improvement of $2.6\%$, which is due mostly to a much higher recall. 

Note that the method by Poira et al. \shortcite{cambria2016} uses also features extracted from other datasets for sentiment, emotion and personality classification, as these features are considered to be useful for the task of sarcasm detection. 
Moreover, as our goal is to propose a baseline, the training time in the order of minutes is an advantage of our model.
We report such results as an upper bound considering that our model does not use additional information from external data. 

The best results are obtained using the star labels. In this setting, our best-performing classifiers are better than the $74.4$ F-score value reported by Buschmeier, and our best $F$-score of $80.7$ represents an $8.5\%$ relative improvement. In this single case of SarcasmCorpus*, the results with the Traditional LSA are all higher than their counterparts with Statistical LSA.

\begin{figure*}[t]
	\centering
	\includegraphics[scale=0.53]{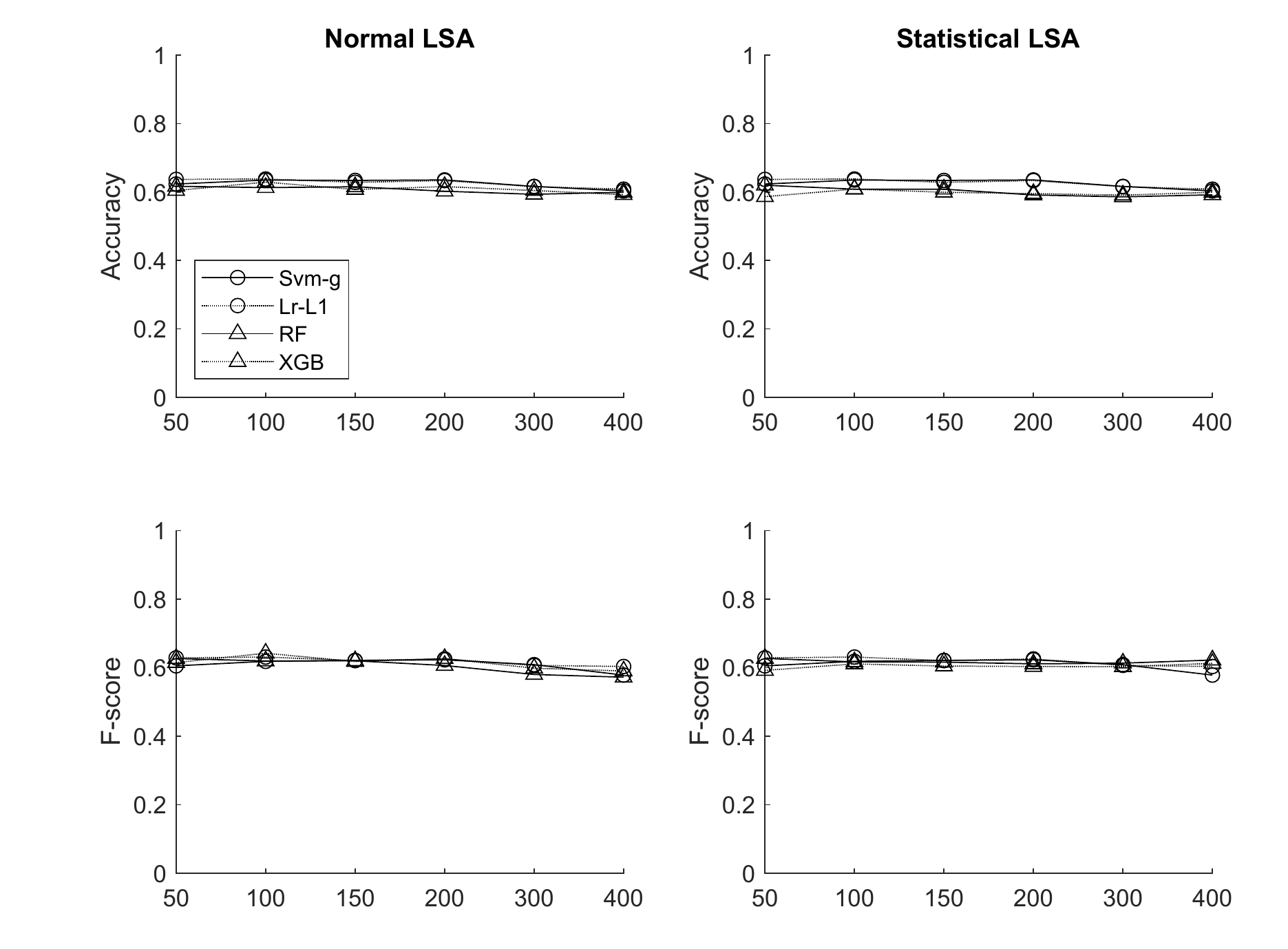}
	\caption{Accuracy and F-score values of all the considered classifiers for different LSA size in the case of IAC-Sarcastic Corpus}
	\label{fig:IAC-Sarcastic}
\end{figure*}

\subsubsection{IAC-Sarcastic}
For IAC-Sarcastic we do not have any previously published result to compare with. The only related result is reported in Joshi et al. \shortcite{joshi-sharma-bhattacharyya:2015:ACL-IJCNLP}, which uses a corpus randomly extracted from IAC containing 752 sarcastic and 752 not sarcastic texts. They report an F-score of $64.0$ (average over a 5-fold), but the text sampling procedure is not specified in the paper. Thus, we prefer to use the sarcastic selection given by the Internet Argument Corpus website\footnote{https://nlds.soe.ucsc.edu/sarcasm1} which is also a bit larger (998 sarcastic and 997 non-sarcastic texts).\\
Accuracies and F-scores of all the classifiers at varying T-SVD size are plotted in Figure \ref{fig:IAC-Sarcastic}, best values of F-score, precision and recall are reported in column IAC-Sarcastic of Table \ref{main-table2}. The best result (F=$66.8$) is lower than in SarcasmCorpus, despite IAC-Sarcastic being balanced and larger than SarcasmCorpus.
With Traditional LSA the $F$-scores are generally slightly lower, but the precision values are higher. \\
The results from Poira et al. \shortcite{cambria2016} are significantly higher, suggesting that in this dataset the sarcasm can be detected in most cases with the linguistic features used by their network independently from the context. 

\begin{figure*}[t]
	\centering
	\includegraphics[scale=0.53]{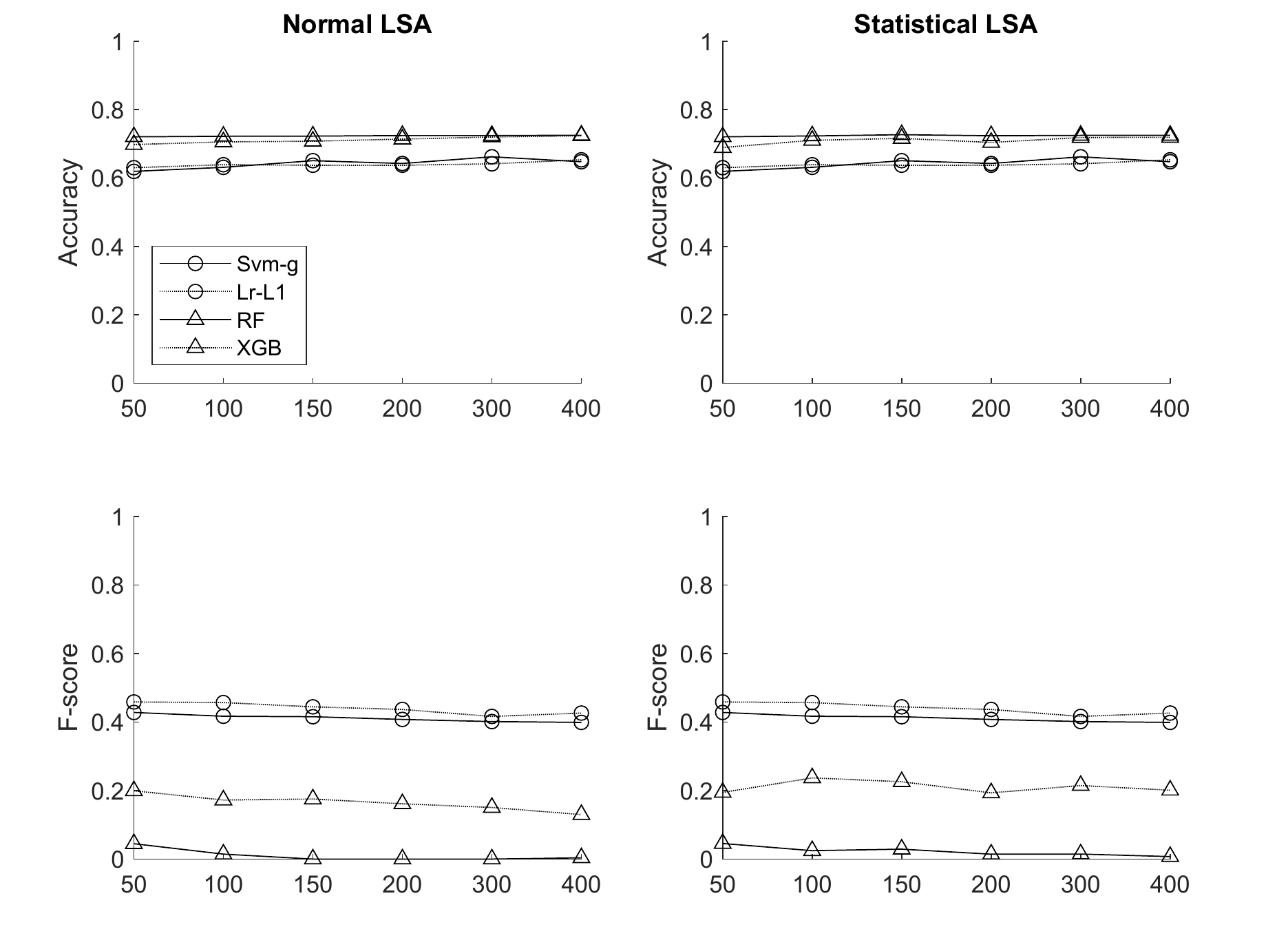}
	\caption{Accuracy and F-score values of all the considered classifiers for different LSA size in the case of Irony-context Corpus}
	\label{fig:Irony-Context}
\end{figure*}

\subsubsection{Irony-Context}
For the irony-context corpus, we used the same 1949 documents selected for the experiments reported in Wallace et al. \shortcite{wallace2014humans}. To allow fair comparisons, we used only the texts of the comments, without any contextual information. \\
The authors report a mean F-score over the five-fold of 0.383 by using a bag-of-words representation with $50,000$ tokens, plus some other binary features that have proven useful in other works, and an SVM classifier with a linear kernel. Our results are plotted in Figure \ref{fig:Irony-Context} and reported in column irony-context of Table \ref{main-table2}, where it is shown how our classifiers clearly outperform the baseline. Our maximum F-score of $46.0$ represents a relative improvement of 20\%. 
Moreover, it is important to highlight the incredibly low values obtained in this corpus when compared with the results from the previous corpora. This is certainly due to the high skewness between the classes; in fact, the positive samples are just 537 over 1949 (27.5\%). If we consider that in SarcasmCorpus the sarcastic texts are only 33\% of the total, we suppose there are other causes. Another reason that can explain the poor results can be found in the diversity of topics, as the texts are extracted from six different forums, and the words used for sarcasm can be highly specific to a given context, both cultural and topical. In Wallace et al. \shortcite{wallace2014humans} it is explicitly said that the request of the context from an annotator is high for the sarcastic texts. As a consequence, classifying correctly the texts without a context is difficult even for humans.
Moreover, the forums from which the posts were extracted are highly controversial, as they regard politics or religion. As a consequence, it is difficult to grasp the sarcasm of a text without knowing the author's opinions. \\
The results with Traditional LSA are very similar to Statistical LSA, and the real surprise is the incredibly low scores obtained by the random forest and gradient boosting methods. \\

\begin{figure*}[t]
	\centering
	\includegraphics[scale=0.53]{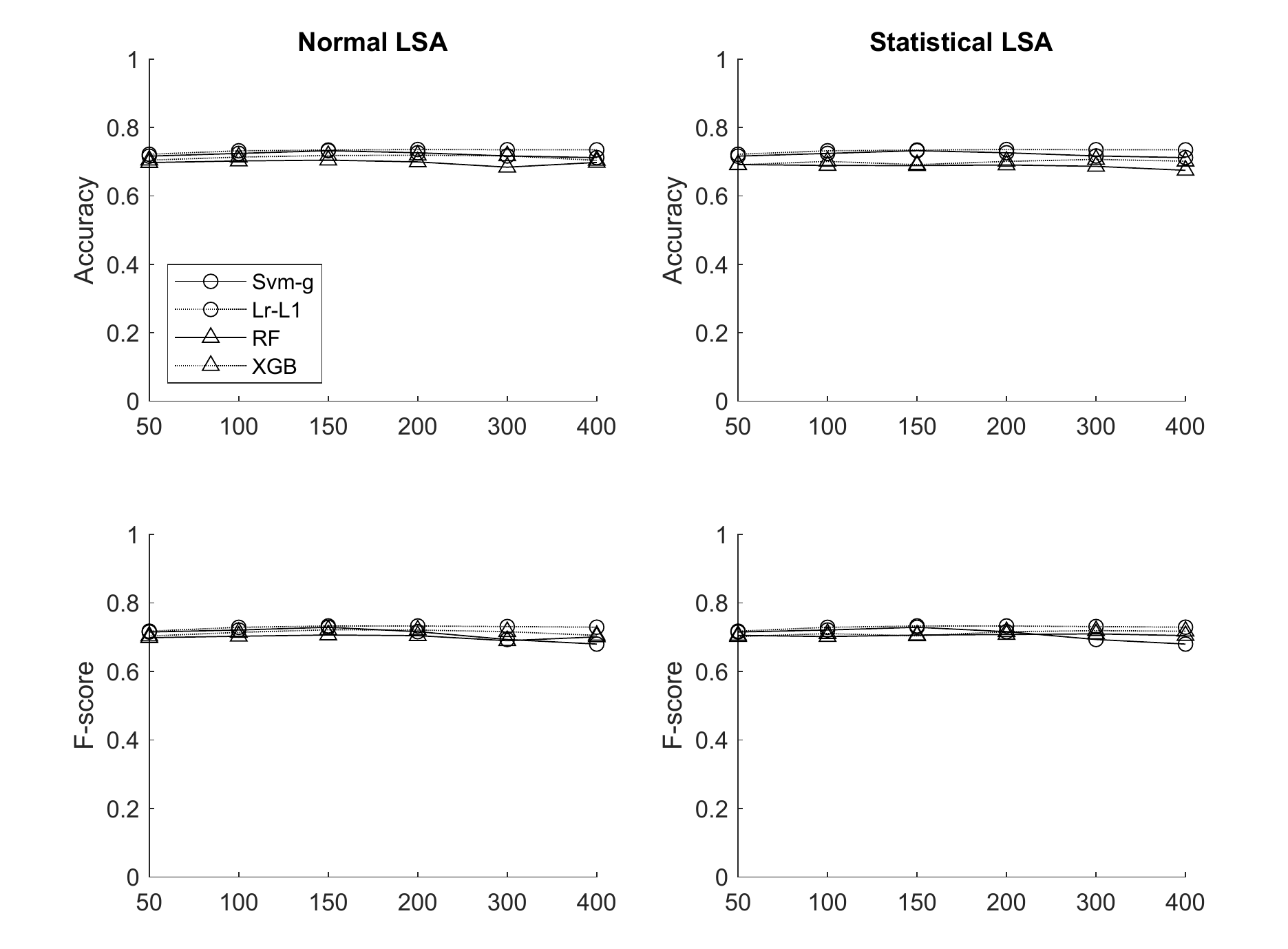}
	\caption{Accuracy and F-score values of all the considered classifiers for different LSA size in the case of IAC-Sarcastic-v2-GEN Corpus}
	\label{fig:IAC-Sarcastic-v2-GEN}
\end{figure*}

\begin{figure*}[t]
	\centering
	\includegraphics[scale=0.53]{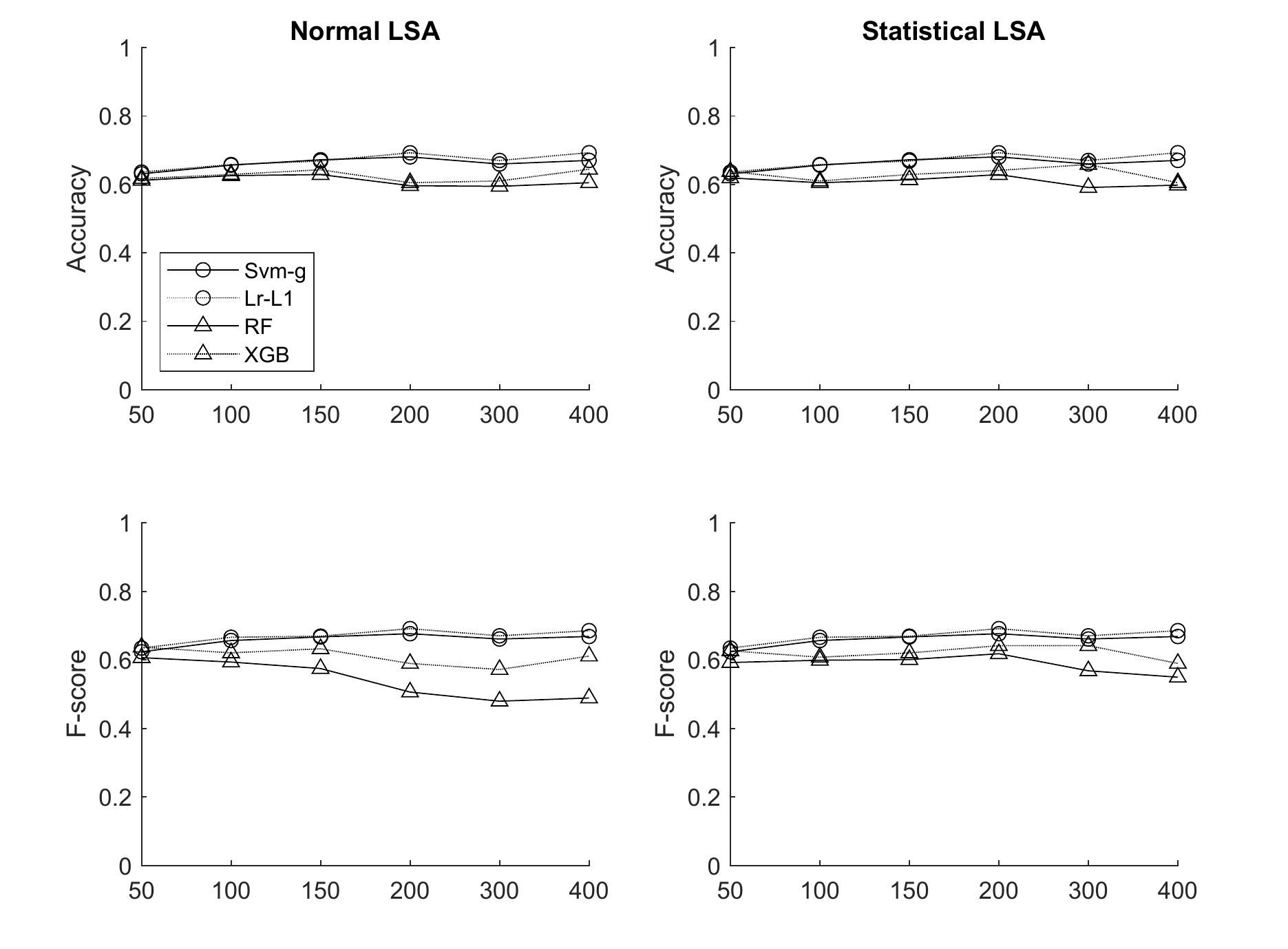}
	\caption{Accuracy and F-score values of all the considered classifiers for different LSA size in the case of IAC-Sarcastic-v2-HYP Corpus}
	\label{fig:IAC-Sarcastic-v2-HYP}
\end{figure*}

\begin{figure*}[t]
	\centering
	\includegraphics[scale=0.53]{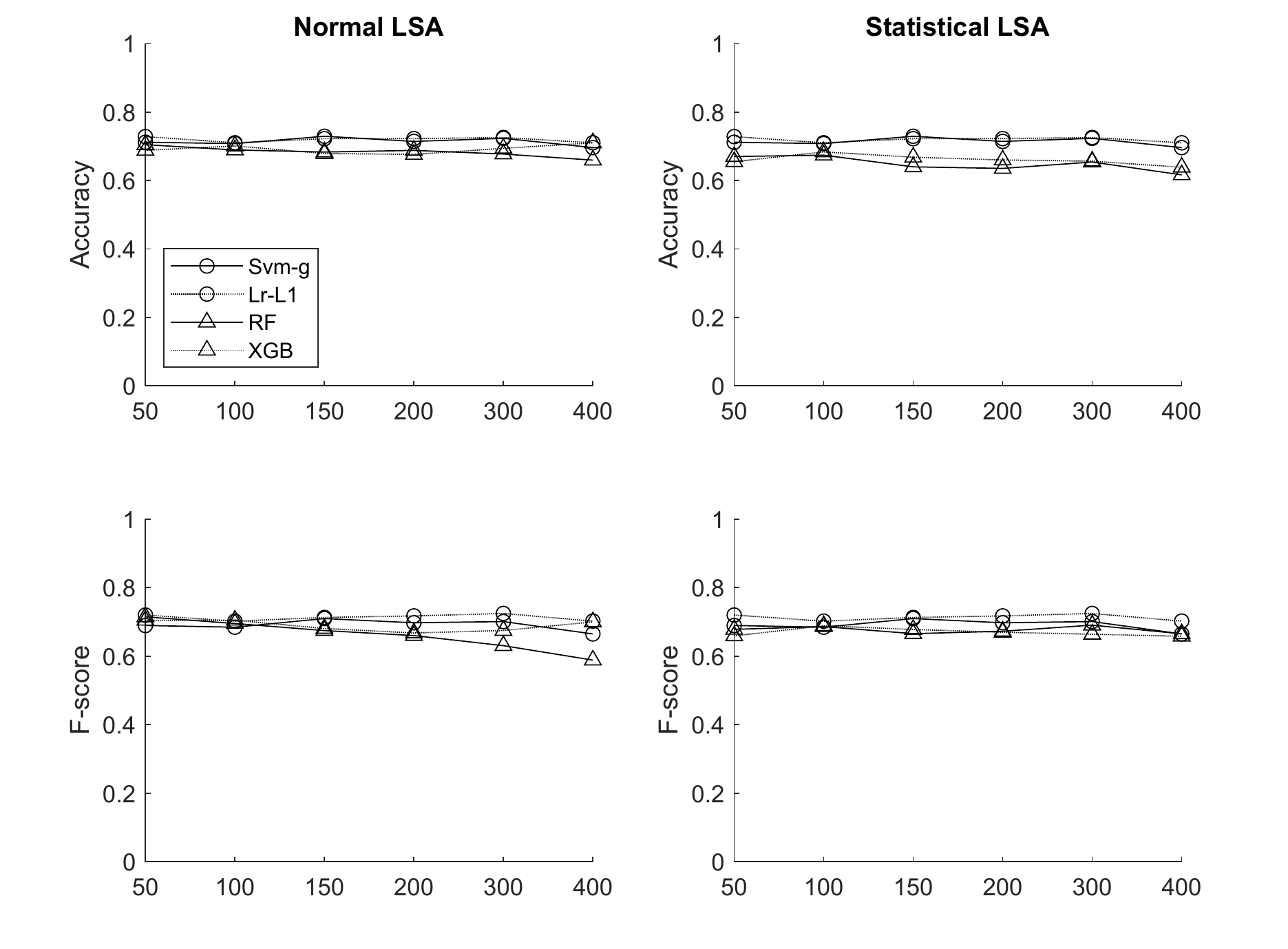}
	\caption{Accuracy and F-score values of all the considered classifiers for different LSA size in the case of IAC-Sarcastic-v2-RQ Corpus}
	\label{fig:IAC-Sarcastic-v2-RQ}
\end{figure*}

\subsubsection{IAC-Sarcastic-v2}
In this case, we wanted to compare our results against those from Oraby et al. \shortcite{oraby2016creating}, which deal with the three sub-corpora separately. However, they are not directly comparable because at the moment in which we report these results only half of the corpus has been released, consisting of $3260$ posts in the generic sub-corpus, $582$ in the Hyperbole side and $850$ for rhetorical questions. The three sub-corpora are all balanced.\\
Results computed on the three subcorpora are plotted in Figures \ref{fig:IAC-Sarcastic-v2-GEN}, \ref{fig:IAC-Sarcastic-v2-HYP}, \ref{fig:IAC-Sarcastic-v2-RQ} and reported in the last three columns of table \ref{main-table3}. 
Despite the difference in data availability, the results are quite encouraging. In fact, we can see that our method reaches the $F$- score of $74.9$ in the generic sub-corpus, slightly better than the previous study. 
Moreover, it improves over Oraby et al. (2016) also in the other two sub-corpora but using Traditional LSA.\\
Nonetheless, these results show that it is possible to achieve very good performance when high-quality labeled corpora are available, even with a limited number of examples. \\
For the CNN, we have results only in the generic sub-corpus, and this is the only case in which at least one of our models can outperform it in terms of F-score.

\begin{table}
\begin{center}
	\caption{Results on the dataset provided for the shared task on irony detection in tweets SemEval2018 Task 3A. The first $11$ rows report the Accuracy (acc), precision (prec), recall (rec) and $F_1$ score ($F_1$) of all the submissions. The last four rows report the results we have performed on the dataset regarding a list of proposed classifiers.}
	\begin{minipage}{\textwidth}
		\begin{tabular}{ccccc}
		\hline\hline
		Method & acc   & prec   & rec & F$_1$\\
		\hline
        THU NGN - Wu et Al. \shortcite{wu2018thu}  &\textbf{73.5} &63.0 &80.1 &\textbf{70.5} \\
 		NTUA-SLP - Baziotis et Al.\shortcite{baziotis2018ntua} &73.2 &\textbf{65.4} &69.1 &67.2 \\
		WLV - Rohanian et Al. \shortcite{rohanian2018wlv} &64.3 &53.2 &\textbf{83.6} &65.0 \\
		NLPRLIITBHU - Rangwani et Al. \shortcite{rangwani2018nlprl}&66.1 &55.1 &78.8 &64.8 \\
		NIHRIO - Vu et Al. \shortcite{vu2018nihrio}&70.2 &60.9 &69.1 &64.8 \\
		DLUTNLP-1 &62.8 &52.0 &79.7 &62.9 \\
		ELiRF-UPV - Gonzalez et Al. \shortcite{gonzalez2018elirf}&61.1 &50.6 &83.3 &62.9 \\
		liangxh16 &65.9 &55.5 &71.4 &62.5 \\
		CJ        &66.7 &56.5 &69.5 &62.3 \\
		\#NonDicevoSulSerio - Pamungkas et Al. \shortcite{pamungkas2018nondicevo} &67.9 &58.3 &66.6 &62.2 \\
        SVM BoW Baseline  &63.5 &53.2 &65.9 &58.9  \\
        \hline
		S-SVM      &59.6          &49.4        &76.8    &60.1  \\
		S-Log.Reg.    &62.6 &52.2        &68.2   &59.1 \\             
		S-XGB     &65.1          & 54.6       & 70.7   &61.6  \\
		S-RF     &65.3          & 54.5       & 75.2   &63.2  \\
		\hline\hline
		\end{tabular}
		\vspace{-2\baselineskip}
	\end{minipage}
	\label{tweettable}
	\end{center}
\end{table}

\subsection{SemEval 2018 Task 3A}
The last experiment on a single dataset was performed on the settings of SemEval 2018 Task 3A (Van Hee et al. 2018), which is a shared task on a binary classification of irony, which we introduced in Section~\ref{tweet-dataset}.\\
We start by performing 10-fold cross-validation with our classifiers over varying LSA dimensionality to choose the best setting. 
We used the same set of hyper-parameters used for the previous experiments. \\
Once we have found the best setting, we train again the model with all the data and predict the classes of the test tweets. We found that we obtain the best results in cross-validation with LSA vectors of size $20$, and the results are presented in Table \ref{tweettable}. We list results for four different classifiers, namely logistic regression, support vector machine, gradient boosting and random forest. In this case, we get the best results using random forest, followed by gradient boosting. In particular, random forest obtains an F$_1$-score of $63.2$, which is higher than the 6th submission.
It is worth noting that the submissions that we listed in the Table, except for the baseline, all use approaches based on deep learning.
Compared to the unigram SVM baseline used for the shared task (row $11$ in table $4$), our model with the random forest is clearly better according to all the metrics, while our model with SVM is better in terms of F$_1$ score but not accuracy.\\ 
Clearly the model we provide is not the best one in terms of accuracy, and showing its superiority among all the others does not represent the goal of this work, but the best performers, i.e. deep learning networks, involve a high number of parameters and high computational training cost. Moreover, there are additional interesting notes. First, the submission by \cite{gonzalez2018elirf} also makes use of deep neural networks but does not get a higher score than our best. Second, the submission by \cite{pamungkas2018nondicevo} is using SVMs over syntactic, semantic, and affective features, but still is not better than our best score. The models that showed a clear superiority use deep networks pre-trained on external data to extract more meaningful features. Thus, while the advantage is real, the number of parameters and the amount of data used is much higher.
\\

\subsection{Inter-corpora Experiments}
The second group of experiments is aimed at finding whether the sarcasm is domain-dependent, or the knowledge acquired over one dataset can be transferred to another.
We evaluate the similarity among the datasets by training a model over all the data of a corpus and using a second corpus as a test set. 
Our best results for every corpus pair are listed in Tables \ref{tabletestSarcasmIAC} and \ref{tabletestironyIACv2}, where the rows indicate the training set and the columns the test set.
Quite interestingly, unlike the in-corpus experiments where the logistic regression works better in some cases, all the top scores that we report for these experiments are obtained by using the SVM classifier.

In Table \ref{tabletestSarcasmIAC} we find the results for SarcasmCorpus and IAC-Sarcastic used as test sets. 
For the case of SarcasmCorpus, the F-scores are quite low compared to the in-corpus experiments. In fact, here we obtain the best result of only $48.5$ when IAC-Sarcastic is the training set, which is much lower than the scores of about $70$ that we get in the in-corpus experiments (column SarcasmCorpus in table \ref{main-table1}). The low results suggests to us that the sarcasm conveyed by the texts in SarcasmCorpus is somehow different from what we can observe in the other corpora. 

When we use IAC-Sarcastic as a test set, we can observe higher scores (column IAC-Sarcastic in table \ref{tabletestSarcasmIAC}), and the F-score of $67.2$ that we obtain by training in IAC-Sarcastic-v2 is comparable to the $66.8$, which is the best result in the in-corpus experiments. Also, the lower result, which we obtain when training on irony-context, is quite close to the result obtained for the in-corpus experiment, and unexpected since the poor results obtained in the in-corpus experiments for irony-context (column Irony-Context in table \ref{main-table2}). 

When irony-context is the test set (first three columns of table~\ref{tabletestironyIACv2}), we can observe again that the F-score obtained by training in IAC-Sarcastic-v2 is higher than the score obtained in the in-corpus experiment.  Nonetheless, all the scores for this test set are lower than $50\%$ with high recalls and low precisions. 

When using IAC-Sarcastic-v2 as the test set (see last three columns of Table \ref{tabletestironyIACv2}) we can observe F-scores between $66.5$ and $70.8$ and are characterized by a high recall and lower precision. The top F1 score is obtained when using IAC-Sarcastic as a training set, which also corresponds to the highest precision. This represents a further proof in favor of the similarity of the two corpora. The top recall score of $91.4$ is obtained by training on SarcasmCorpus, but the precision is much lower than the other two cases.

Overall, it is worth noting that, for all the experiments, the top results are obtained by training on either IAC-Sarcastic or IAC-Sarcastic-v2, while SarcasmCorpus is always better than irony-context. Considering that the quality of the features depends on the quality of the data and of the annotation, we suppose that the quality of the first two datasets is higher than the quality of irony-context, while the data contained in SarcasmCorpus are too different from the other corpora. A deeper analysis of the corpora can be found in the discussion (Section \ref{discussion}).
{\small
	\begin{table}
		\caption{Best inter-corpora results from every dataset to SarcasmCorpus and IAC-Sarcastic. The results were selected according to the F$_1$ score. Rows indicate the used training set, while column the test set.}
		\begin{minipage}{\textwidth}
			\begin{tabular}{lcccccc}
				\hline\hline
				& \multicolumn{3}{c}{SarcasmCorpus} & \multicolumn{3}{c}{IAC-Sarcastic}\\ 	
				\hline
				 & F$_1$ & prec & rec & F$_1$ & prec & rec\\
				\hline
				SarcasmCorpus &- &- &- &65.8& 52.4 & 88.3 \\ 	
				IAC-Sarcastic &48.5 &39.7 &62.2 &- &- &-\\ 
				irony-context &43.5 &46.6 &40.7 &64.2 &55.9 &75.4\\ 
				IAC-Sarcastic-v2 &47.2      & 43.7      & 51.3 &67.2 &61.9 &73.4\\ 	
				\hline\hline
			\end{tabular}
			\vspace{-2\baselineskip}
		\end{minipage}
		\label{tabletestSarcasmIAC}
	\end{table}
}

{\small
	\begin{table}
		\caption{Best inter-corpora results from every dataset to irony-context and IACv2. The results were selected according to the F$_1$ score.}
		\begin{minipage}{\textwidth}
			\begin{tabular}{lcccccc}
				\hline\hline
				& \multicolumn{3}{c}{irony-context} & \multicolumn{3}{c}{IAC-Sarcastic-v2}\\ 	
				\hline
				 & F$_1$        & prec      & rec       & F$_1$        & prec      & rec\\
				\hline
				SarcasmCorpus   & 44.1      & 30.0      & 83.6      & 67.2      & 53.2      & 91.4\\ 	
				IAC-Sarcastic  & 45.4      & 32.0      & 78.0      & 70.8      & 60.4      & 85.4\\ 
				irony-context &- &- &-   & 66.5      & 58.0      & 77.9\\ 
				IAC-Sarcastic-v2 & 47.8      & 35.4      & 73.6 &- &- &- \\ 				\hline\hline
			\end{tabular}
			\vspace{-2\baselineskip}
		\end{minipage}
		\label{tabletestironyIACv2}
	\end{table}
}

\begin{table}
	\begin{center}
	\caption{Results of the union experiments on SarcasmCorpus and IAC-Sarcastic. With a 10-fold, each model has been trained on the concatenation of 9 folds of each corpus and tested on the remaining fold.}
	\begin{minipage}{\textwidth}
		\begin{tabular}{ccccccc}
		\hline\hline
		& \multicolumn{3}{c}{SarcasmCorpus} & \multicolumn{3}{c}{IAC-Sarcastic}\\
		\hline
		Method & F   & prec   & rec & F   & prec   & rec\\
		\hline
		S-SVM gauss    &62.1 &63.3 &61.6 &66.7 &60.5 &75.2\\
		S-Log. Reg.L1  &62.1 &66.7 &58.1 &65.4 &63.8 &67.1\\
        S-RF     	   &26.0 &78.8 &15.8 &66.1 &77.7 &57.6\\            
        S-XGB     	   &47.8 &69.1 &37.3 &68.7 &76.7 &62.3\\
		\hline\hline
		\end{tabular}
		\vspace{-2\baselineskip}
	\end{minipage}
	\label{table:union-experiment-1}
	\end{center}
\end{table}

\begin{table}
	\begin{center}
	\caption{Results of the union experiments on irony-context and IAC-v2. With a 10-fold, each model has been trained on the concatenation of 9 folds of each corpus and tested on the remaining fold.}
	\begin{minipage}{\textwidth}
		\begin{tabular}{ccccccc}
		\hline\hline
		&  \multicolumn{3}{c}{irony-context} & \multicolumn{3}{c}{IAC-v2} \\
		\hline
		Method         &F     &prec  &rec   &F    &prec &rec\\
		\hline
		S-SVM gauss    &46.5  &35.9  &66.3  &71.8 &63.4 &82.8\\
		S-Log. Reg.L1  &44.5  &36.6  &54.9  &71.0 &67.0 &75.5\\
        S-RF      	   &40.3  &40.8  &40.0  &68.8 &65.2 &63.5\\
        S-XGB     	   &41.9  &40.2  &44.2  &72.7 &74.6 &70.9\\
		\hline\hline
		\end{tabular}
		\vspace{-2\baselineskip}
	\end{minipage}
	\label{table:union-experiment-2}
	\end{center}
\end{table}

\begin{table}
\caption{Some examples from dataset SarcasmCorpus. The first three examples are sarcastic and the following three are not sarcastic.}
\begin{tabular}{l}
\hline\hline
one word: punctuation. ok, folks heres the deal: if you enjoy reading books where \\ 
there are NO QUOTATION MARKS and punctuation such as that, then go \\	ahead, be my guest. if you happen to be SANE, then you would agree with me \\
when i say this: it is thuroughly impossible to read a book when you cant tell who \\ 
the heck is talking. what idiot translated this?! oh, and if you happen to like other \\
things too, like, say, a good, interesting, or mildly understandable plot, then youre \\
out of luck. god, the people around here. \\
\noalign{\vspace {.5cm}}
Is there somebody you really don't like, want to torture perhaps?  Give them a ticket\\
to see this movie, or a copy of the forthcoming DVD (which in this case stands for \\ 
Dreadful, Very Dreadful). Take 5\% Prophecy, 75\% Night of the Living Dead, 20\% \\ 
Bad LSD, stir well with pitiful writing, and VOILA! You have Legion.
 The old lady \\
in the diner scene might have been lifted from IT, the rest of the movie was lifted \\
from something that rhymes with IT. Whoever really wrote this garbage should be \\
assigned to the ninth circle of the inferno.  The big guy doesn't like us anymore?  \\
No wonder, he's got an English accent in this movie, as do his angels.  Perhaps they\\
are still sulking about the revolutionary war. Don't watch this drivel. \\
\noalign{\vspace {.5cm}} 
but thought it was to funny, this book is for you. A book about nothing, that takes \\ 
200 plus pages to get there. If you chose to read it keep sharp objects out of reach, \\
because you may feel like poking your eyes out. \\
\hline 
I have used Olympus cameras in the past without any complaints.  This camera \\
holds up to our expectations.  The picture quality is amazing.  The scene modes \\ 
perform well, but we don't use them that often, we prefer to point and shoot with \\
the auto-mode. This camera is replacing a 3MP nikon coolpix, that finally wore out.  \\
Just remember when you buy this camera, it is a pocket sized digital camera for \\
around \$100. Don't expect to be taking photos for National Geographic!!  It is \\ 
a great little camera  :) \\
\noalign{\vspace {.5cm}} 
This is controller works like a charm. You save money and time by getting it from \\
amazon too. All in all buy this. \\
\noalign{\vspace {.5cm}} 
This is a great travel tray for my 3 year old son.  We have a Britax carseat \\
and this fits perfectly.  It is just sturdy enough that he is able to eat snacks, \\
color, and play with his cars and other small toys.  It also keeps everything \\
from constantly falling on the floor.  The mesh pockets on the side are also a \\
great feature. \\
\hline\hline
\end{tabular}
\label{table:filatova-examples}
\end{table}

\begin{table}
\caption{Some examples from IAC-Sarcastic. The first three examples are sarcastic and the following three are not sarcastic.}
\begin{tabular}{l}
\hline\hline
Where'd Noah get Tasmanian tigers? \\
\noalign{\vspace {.5cm}} 
Isn't it interesting that TQ is so hard up for a reason to attack me that he must \\
himself use a fallacious position to do it from. Thanks TQ, for being so consistent\\
in your dishonesty. \\
\noalign{\vspace {.5cm}} 
''No, actually the earth is 150 years old. FACT. And its age never changes. FACT.''\\
-Chris Formage, Grand Theft Auto: San Andreas. \\
\hline
But you think SETI is worthwhile! \\
\noalign{\vspace {.5cm}} 
So why are you here? Do you have anything to offer of any substance? \\
\noalign{\vspace {.5cm}} 
To the extent that working toward social change on any moralistic issue is an \\
'imposition', sure. But I think one could say that about anything that has a \\
social effect. \\
\hline\hline
\end{tabular}
\label{table:iac-examples}
\end{table}

\begin{table}
\caption{Some examples from irony-context. The first three examples are sarcastic and the following three are not sarcastic.}
\begin{tabular}{l}
\hline\hline
Democrats don't know how to manage money?   Shocking! \\ 
\noalign{\vspace {.5cm}} 
I'm sure it will pass.. \\
\noalign{\vspace {.5cm}} 
That was today's talking point. \\ \hline

Would he win? \\
\noalign{\vspace {.5cm}} 
Yeah I didn't get far.  This article fills me with sadness \\
\noalign{\vspace {.5cm}} 
I saw it best said in another cartoon "At no time was the Obama administration \\ 
aware of what the Obama administration was doing."" \\ 
\hline\hline
\end{tabular}
\label{table:irony-examples}
\end{table}

\begin{table}
\caption{Some examples from IAC Sarcastic v2. The first three examples are sarcastic and the following three are not sarcastic.}
\begin{tabular}{l}
\hline\hline
"More chattering from the peanut gallery? Haven't gotten the memo, you're no \\
longer a player? Honestly....clamoring for attention is so low budget. No shame." \\
\noalign{\vspace {.5cm}}
And I would appreciate your refraining from calling your mythology a theory. \\
\noalign{\vspace {.5cm}}
"yeah just like you and your idols Glenn Beck, Rush Limbaugh, etc."\\
\hline
"You mean that it must have has a beginning, don't you?" \\
\noalign{\vspace {.5cm}}
"Yes, but if seen from space (something the ancients couldn't do) it LOOKS \\
like it hangs on nothing. Don't worry, I don't mind spelling it out for you." \\
\noalign{\vspace {.5cm}}
You lost me there. Try them how and for what? \\
\hline\hline
\end{tabular}
\label{table:sarcasmv2-examples}
\end{table}

\subsection{Union Experiments}
The last group of experiments we ran has the goal of understanding whether the combination of data coming from different sources can influence positively the final score. 
For this purpose, as anticipated in Section~\ref{sec:exp_setup}, we computed 10-folds of each of the four corpora used for the first group of experiments, and used as a training set the concatenation of $9$ folds of every corpus, and as a validation set the remaining single folds of each corpus.

From Tables \ref{table:union-experiment-1},\ref{table:union-experiment-2} we can observe that these results are not higher overall with respect to the inter-corpora results. The only exceptions are SarcasmCorpus, where the results are almost $20$ F-score points higher than those obtained in the inter-corpora; and IAC-v2, where the gradient boosting (XGB) obtains $2$ F-score points more than the top score in the inter-corpora results. 

The results on SarcasmCorpus are still lower than the in-corpus results, and the scores of random forest and gradient boosting are much lower than the other two methods. This is further evidence that adding diverse data is not helpful, or is actually harmful, for classifying SarcasmCorpus. \\
The general trend of this block of experiments is that our classifiers are not able to leverage data from different domains in order to improve global results. In-domain data represent the best choice even if the data amount is lower.

\subsection{Discussion}
\label{discussion}
In this section, we discuss our results from a more general point of view. We start by briefly discussing the content of the different corpora. Then we try to relate the results of the different types of experiments. Finally, we detect the limits of our experiments for the type of documents we worked with.

The corpora we used for our experiments are characterized by high internal variability in style, as each corpus consists of texts from thousands of different authors. 
Despite the number of authors, there are some factors that depend on the type of text and the medium. For instance, the irony-context, IAC Sarcastic, and IAC Sarcastic v2 corpora are made of posts collected from online forums, which are mostly about politics. Most of the texts are extracted from longer arguments, and thus the style is informal and in general with aggressive tones. 

In Tables \ref{table:iac-examples}, \ref{table:irony-examples} and \ref{table:sarcasmv2-examples} we show some randomly selected samples from these corpora. As is apparent from the samples, the posts have a target to attack, who can be another user or the subject of the discussion. 
Table~\ref{table:iac-examples} shows some examples from IAC-Sarcastic. In all the examples the author attacks another user or their opinions. For instance, the first and the third sarcastic examples make sarcasm about the Bible to attack another user's religious ideas, while in the second example the author uses sarcasm to expose a fallacious position of another user and not appearing rude on his side. By contrast, the non-sarcastic examples are much more direct about their meaning.
A similar pattern can be found in the examples from IAC Sarcastic v2 (table~\ref{table:sarcasmv2-examples}). Sarcasm is again used to attack a person (first example) or his/her opinions (second example), maybe religious. The third example shows that also in this corpus some sentences are hard to classify. In this case, the information that we get is that the target has ultraconservative ideas, but it is not easy to grasp the sarcasm.
The examples from irony-context (in table~\ref{table:irony-examples}) are much more difficult to grasp without knowing contextual information. For instance, the first sarcastic example can be either sarcastic or literal according to the political opinion of the author. It is sarcastic if the author is a Republican, it is not sarcastic (but would appear strange to write) if the author is a Democrat. The second and the third examples are hard to classify without knowing the subject of the conversation. The same issue of missing a broader context also appears in the non-sarcastic examples, and the third examples can easily be interpreted as sarcastic by humans.
In SarcasmCorpus the situation is different as there is no argument ongoing, and the sarcasm is made against products that the author did not like.
In this case, there are many references to the external world and the writing is more vehement in its negative stance. Some samples are shown in Table \ref{table:filatova-examples}.
The sarcastic examples in table \ref{table:filatova-examples} all express a negative sentiment and also use negative words. Sarcasm is used within this negative reviews to attack the product in a more creative way and make the text more fun than a usual negative review. The non-sarcastic reviews, on the other side, give a description of the product and their experience with it, with literal forms of expressing the sentiment (``are also a great feature'', ``It is a great little camera'').
We suppose that this difference in style is the main obstacle to correct classification of SarcasmCorpus instances in the cross-corpora experiments.\\
We now discuss the relations among the results of the different experiments to gain some further insights into the sarcastic content of our corpora.
From the in-corpus experiments, we obtain good results on SarcasmCorpus, which is the only corpus containing Amazon reviews. Unfortunately, when we train our models in a cross-corpora or all-corpora setting, our results drop dramatically, especially in the cross-corpora case. These results mean that the sarcasm in SarcasmCorpus is conveyed through features that are not present in the other corpora. This is especially true when considering that in the inter-corpora experiments, using SarcasmCorpus as a training set in all cases yields results that are only better than the ones obtained when using irony-context as a training set.

The results on irony-context show that this corpus is much more difficult to classify than the others, as it was pointed out also in the paper that presented it (Wallace et al. 2014), which highlights how the human annotators needed to read the contexts to be sure about the sarcastic posts. 
In the inter-corpora experiments, the results when training on irony-context are the worst for all the test sets, but only for a few points of F-score, while at first, we could expect dramatically lower results. For us, the previous are strong suggestions that the types of texts present in irony-context are similar to the ones present in IAC-Sarcastic-v2, but the quality is lower. As a consequence, this is a further proof that the dataset annotators do not consider sarcasm and irony to be two different linguistic phenomena.

The two versions of IAC-Sarcastic have proved to be the easiest to classify when using other corpora for training.
The best result in IAC-Sarcastic is obtained in the Union experiment (see Tables \ref{table:union-experiment-1},\ref{table:union-experiment-2}), and thus it benefits from the higher amount of data, especially from the data from IAC-Sarcastic-v2, as can be observed from the cross-corpora results (Table \ref{tabletestSarcasmIAC}).

By contrast, the best results on IAC-Sarcastic-v2 are obtained with the in-corpus experiments, while all the results obtained in the inter-corpora experiments are clearly worse. Among the inter-corpora, training the model with IAC-Sarcastic results in a F\_score of $70.8$, which means a relative decrement of $5.4\%$ concerning the top score for the intra-corpus experiments of IAC-Sarcastic-v2. It is interesting to note that one cause of the decrement can also be the size of corpora, in fact, IAC-Sarcastic contains only $1995$ texts, while IAC-Sarcastic-v2 contains $3260$.  \\
One final remark is about the absolute scores obtained in the in-corpus experiments. In fact, we can notice that in SarcasmCorpus the F\_score can go beyond $0.7$, and up to $0.8$ by adding the star rating as a feature. The high result can be explained by the peculiarity of this corpus, where sarcasm is present mostly in negative reviews, and the star label is the single best indicator of sarcasm \cite{buschmeier2014impact}. The other corpora consist of texts that belong to a thread of forum posts. Sometimes it is reasonable to classify such posts as sarcastic or not out of context, but in many cases, it is impossible also for humans (see examples in Table \ref{table:irony-examples}). In fact, the low F\_score in irony-context is due to low precision, which is an indicator of high similarity between the positive and negative classes. Moreover, low precision and higher recall is a pattern that is present in most of the experiments, even if with higher absolute numbers. The combination of high recall and lower precision suggests that the dubious texts are classified as sarcastic more often than not sarcastic.

\section{Conclusions}
\label{conclusion}
In this work, we have tackled the problem of automatic sarcasm detection from a data-driven point of view. In more detail, we have used a set of labeled datasets and applied distributional semantics followed by some machine learning approaches in order to give a baseline for the literature in managing such a problem.
We do not differentiate between sarcasm and irony because they are not so easily distinguishable even for human experts.
Experiments have been carried out on four different corpora containing texts from online reviews or forums, and the corpus used for the shared task on irony detection on Twitter proposed in SemEval 2018.
We have shown experimentally that some basic methods can outperform in all the datasets other methods based on bag of words and linguistic features, thus representing a solid baseline. 
With our experiments that train the models with one corpus and test them by using the other corpora, we have confirmed experimentally that also the annotators tend to not distinguish the distinction between irony and sarcasm. By contrast, major differences can be found according to the text domains, i.e., review vs. political forum. The domain difference can also prevent the method from taking benefits from more data when they are too diverse from the test data.
As a future work, we will try to improve distributional semantics approaches with linguistic features in order to perform more fair comparisons with more recent and advanced methods, also in a cross-lingual scenario, e.g. by leveraging the data from the Evalita shared task on irony detection in Italian tweets\footnote{http://www.di.unito.it/~tutreeb/ironita-evalita18/index.html}. Furthermore, we will exploit more classical AI methodologies (e.g., by using ontologies, reasoners, common-sense reasoning techniques, etc.)  to deduce the context, understanding the concepts expressed in a sentence, exploiting also features like hashtags and emojis to improve the overall performance of the approach.



\label{lastpage}
\end{document}